\PassOptionsToPackage{table}{xcolor}

\documentclass{article} 
\usepackage{iclr2027_conference,times}

\usepackage{hyperref}
\usepackage{url}

\usepackage[T1]{fontenc}
\usepackage[utf8]{inputenc}
\usepackage{microtype}
\usepackage{graphicx}
\usepackage{amsmath}
\usepackage{adjustbox}
\usepackage{xcolor}
\usepackage{float}

\title{When Order Matters: First-Speaker Bias and Mitigation through Personality in Sequential Multi-Agent Debate}

\author{Duofeng Xu \quad Bryan Hooi \quad Dandan Qiao \\
School of Computing \\
National University of Singapore \\
Singapore \\
\texttt{xu.duofeng@u.nus.edu.sg, dcsbhk@nus.edu.sg, qiaodd@nus.edu.sg}
}

\iclrfinalcopy 

\begin{document}

\maketitle
\lhead{}

\begin{abstract}

Multi-agent debate (MAD) is often used to improve large language model (LLM) reasoning, but sequential debate is rarely a neutral aggregator of agents' opinions. We show that sequential MAD suffers from a pronounced \textbf{first-speaker bias}: agents disproportionately shape the final answer when they speak first. As a result, placing a stronger model after weaker ones can substantially offset its reasoning advantage. We then focus on the disadvantaged strong-agent-last setting and ask whether \textbf{personality prompting} can mitigate this imbalance. Drawing on the Big Five model, we study agreeableness and extraversion as behavioral interventions applied to either the strong or weak side. We find that their effects are trait-specific. Influence consistently shifts in the direction of lower agreeableness, and assigning low agreeableness to the stronger agent helps restore its lost influence and improves final accuracy. Extraversion, by contrast, produces less systematic changes in influence and accuracy, with its clearest effect appearing in agents' verbosity. These findings show that effective MAD design depends not only on model capability, but also on how speaking order and induced interaction behavior shape the debate process. \footnote{Code is available at \url{https://anonymous.4open.science/r/personality-MAD-code}}

\end{abstract}

\section{Introduction}

Multi-agent debate (MAD) has emerged as a popular approach for improving large language model (LLM) reasoning, allowing multiple agents to propose, critique, and revise answers through interaction~\citep{pmlr-v235-du24e, liang-etal-2024-encouraging}. A common implementation is sequential debate, in which agents speak one after another in a fixed order~\citep{ICLR2024_25cc3adf}. The protocol is appealingly straightforward to implement, but it implicitly assumes that speaking order does not systematically bias the outcome.

This implicit assumption is consequential. In human teams, decisions often depend less on who has the best information than on who speaks first~\citep{WEISBAND1992352} and who is willing to push back~\citep{SCHULZHARDT2002563}. A junior member who speaks first can anchor a discussion that more senior colleagues never fully recover from; teams without at least one member willing to hold an unpopular view tend to converge prematurely on whatever was said early~\citep{barryBargainerCharacteristicsDistributive1998a}. We find a strikingly similar pattern in sequential LLM debate. Holding the participating models, questions, and protocol fixed and varying only the speaking order, we show that the first speaker exerts disproportionate influence over the final majority answer---strong enough that placing the stronger agent after weaker agents sharply reduces its influence and lowers debate accuracy.

This positional asymmetry suggests a natural intervention. If human teams rely on members who can resist social pressure to keep stronger reasoning alive, can we induce analogous behavior in LLM agents to counteract the disadvantage created when the stronger agent speaks last? We therefore focus our subsequent intervention experiments on the disadvantaged strong-agent-last configuration and examine whether \textbf{personality prompting} can rebalance influence and improve reasoning performance. Specifically, we draw on the Big Five personality model and focus on two of its dimensions, \textbf{agreeableness} and \textbf{extraversion}, because these traits are closely related to interaction behavior in debate~\citep{barryBargainerCharacteristicsDistributive1998a,huang-hadfi-2024-personality}: agreeableness relates to the extent to which an agent accommodates or resists others' views, while extraversion relates to how assertively and expressively it presents its own reasoning. We treat these personality prompts as controlled behavioral interventions and ask whether altering these tendencies can mitigate the strong agent's positional disadvantage and improve collective reasoning.

Concretely, we study three questions:

\textit{RQ1: Does the agent that speaks first exert greater influence over the final answer, and does this affect final reasoning performance of sequential MAD?}

\textit{RQ2: Can agreeableness-based behavioral interventions mitigate the positional disadvantage faced by the strong agent when it speaks last?}

\textit{RQ3: Can extraversion-based behavioral interventions mitigate the positional disadvantage faced by the strong agent when it speaks last?}

Our results suggest that MAD exhibits dynamics that closely parallel those observed in human teams. Specifically, sequential debate exhibits a clear first-speaker bias: the stronger agent exerts substantially less influence, and debate accuracy is lower, when it speaks after rather than before the weak agents. Focusing on this disadvantaged strong-agent-last configuration, we find that personality-guided behavioral interventions can reshape the resulting influence imbalance. Altering agreeableness reliably shifts influence in the direction of lower agreeableness. This in turn affects performance: within the strong-agent-last setting, the largest accuracy improvement occurs when the strong agent is made less agreeable while the weak agents remain unprompted. This pattern suggests that effective debate requires sufficient resistance from the stronger agent without overly suppressing potentially useful contributions from the weak side. Extraversion, by contrast, produces less systematic changes in influence and accuracy; its clearest effect is on how agents express their reasoning, particularly justification length.

Our contributions are twofold. First, we identify first-speaker bias as an important and systematic source of positional influence in sequential MAD, drawing a direct parallel with team-decision dynamics in humans. Second, having identified the disadvantaged configuration, we show that targeted personality-guided behavioral interventions can help mitigate this positional imbalance. In particular, agreeableness provides a controllable mechanism for redistributing influence and, under appropriate configurations, improving final reasoning performance, while extraversion offers a contrasting intervention that primarily affects expressive behavior. Together, these findings show that MAD should be understood not merely as an ensemble of models, but as an interaction process in which capability, speaking order, and agent behavior jointly determine collective outcomes.

\section{Related Work}

\subsection{Multi-Agent Debate for LLM Reasoning}

Multi-agent debate (MAD) has emerged as a prominent framework for improving LLM reasoning by allowing multiple agents to exchange, critique, and revise candidate answers. Prior work has shown benefits across domains, including general reasoning~\citep{pmlr-v235-du24e, liang-etal-2024-encouraging}, mathematical problem solving~\citep{zhang-xiong-2025-debate4math}, and hallucination reduction~\citep{Ma_Gao_Chai_Sun_Wang_Pei_Tao_Song_Liu_Zhang_Cui_2025}. However, these benefits depend strongly on how the debate is structured~\citep{pmlr-v235-smit24a}. Existing studies have explored protocol designs such as confidence-aware debate~\citep{lin-hooi-2025-enhancing}, explicit opposing viewpoints~\citep{fang-etal-2025-counterfactual}, team optimization~\citep{ICLR2024_578e65cd, liu2024a}, specialized roles~\citep{ICLR2024_6507b115, NEURIPS2023_a3621ee9}, judge-based selection~\citep{ICLR2024_25cc3adf, pmlr-v235-khan24a}, and sequential task decomposition~\citep{gu-etal-2025-explain}. These studies show that MAD performance depends both on which models participate and on how their interaction is organized.

\subsection{Interaction Dynamics in LLM Debate}

At the same time, multi-agent collaboration does not guarantee effective deliberation. Several studies show that debate can produce group decision-making failures. Agents may conform to the majority~\citep{zhu-etal-2025-conformity}, copy one another's responses~\citep{pitre-etal-2025-consensagent}, or reach consensus without sufficient independent reasoning~\citep{cisneros-velarde-2025-biases, kaesberg-etal-2025-voting}. Debate may also reinforce existing biases rather than correct them~\citep{NEURIPS2024_32e07a11, oh2025when}, while mixed-capability groups can exhibit asymmetric influence between stronger and weaker agents~\citep{choi-etal-2025-empirical}. These findings suggest that adding more agents does not automatically improve reasoning; debate outcomes are also shaped by influence, dominance, and how agents respond to one another. This issue is particularly important in sequential debate, where each response becomes part of the context for subsequent agents. LLM agents often revise their answers after observing other agents' responses~\citep{cho2025herdbehaviorinvestigatingpeer, ICLR2025_1da9ca7e}, suggesting that interaction history can shape subsequent judgments. Related research on position bias shows that LLM judgments can depend on the ordering of information within a fixed input context~\citep{echterhoff-etal-2024-cognitive}. Recent work further demonstrates that such positional effects can arise within multi-agent debate:~\citet{zhangKeyDecisionMakersMultiAgent2026} vary the positions of independently generated agent viewpoints within the debate update context and find that later-positioned viewpoints exert greater influence on the final outcome, demonstrating that the ordering of presented viewpoints can shape subsequent judgments. Sequential MAD introduces a distinct form of ordering effect: changing the speaking order alters the interaction history itself, because subsequent responses condition on responses generated earlier in the interaction. Speaking order can therefore affect how later reasoning is generated and evolves over the course of the debate. Existing work has not systematically isolated this form of speaking-order effect in sequential MAD. We address this gap by comparing mirrored debate orders while holding the participating models, questions, and debate protocol fixed.

\subsection{Personality in LLM Agents}

Personality prompting provides one way to systematically shape the interaction behavior of LLM agents~\citep{jiang-etal-2024-personallm}. Prior work shows that LLMs can express prompted Big Five traits and that these prompts can produce systematic behavioral changes~\citep{NEURIPS2023_21f7b745,frisch-giulianelli-2024-llm, serapiogarcia2025personalitytraitslargelanguage}. However, it remains unclear whether these behavioral changes can meaningfully affect how agents influence one another during sequential debate, or whether they translate into better reasoning performance. We focus on agreeableness and extraversion of the Big Five personality model because these two traits are especially relevant to interactive debate. Agreeableness relates to cooperation, accommodation, and willingness to accept others' views, while extraversion relates to assertiveness and social expressiveness~\citep{goldbergDevelopmentMarkersBigFive1992a,costaDomainsFacetsHierarchical1995}. These traits are closely linked to how agents argue, respond to disagreement, and update their answers. Both traits have also been linked to negotiation behavior in humans~\citep{barryBargainerCharacteristicsDistributive1998a} and LLM agents~\citep{huang-hadfi-2024-personality}. We therefore use these dimensions as controlled behavioral interventions to examine whether they can systematically reshape influence and performance in sequential MAD.

\section{Experimental Design}
\label{sec:design}
\subsection{Tasks and Evaluation Data} 
\label{sec:data_sets}

We evaluate debate performance on four multiple-choice benchmarks commonly used to assess LLM reasoning: MMLU-Pro, GPQA Diamond, SuperGPQA, and AQuA-RAT. MMLU-Pro~\citep{NEURIPS2024_ad236edc} is a multi-disciplinary benchmark designed to emphasize reasoning. GPQA Diamond~\citep{rein2024gpqa} contains highly challenging graduate-level science questions. SuperGPQA~\citep{NEURIPS2025_a3c5af1f} is a large-scale graduate-level benchmark spanning 285 disciplines. AQuA-RAT consists of algebraic word problems derived from GRE and GMAT exams; we use the version included in AGIEval~\citep{zhong-etal-2024-agieval}. Together, these datasets provide a broad but controlled evaluation setting, and their shared multiple-choice format enables consistent answer extraction and comparison across experiment settings. Across all conditions, we use the same set of questions: 200 randomly sampled questions from each of MMLU-Pro, SuperGPQA, and AQuA-RAT, and all 198 questions from GPQA Diamond.

\subsection{Debate Protocol and Agent Roles}
\label{sec:debate_protocol}

Debate follows a sequential round-robin protocol~\citep{ICLR2024_25cc3adf}. Agents respond in a fixed order within each round, with each response conditioned on all messages available at the time of that agent's turn. Thus, only the first agent’s response in the first round is produced independently; all subsequent responses are conditioned on the accumulated debate history, including the agents’ own prior responses and the responses of other agents. Each agent speaks once per round for four rounds, resulting in twelve utterances per debate. The debate answer is determined by majority vote over the agents' final-round answers. We choose four rounds based on a pilot analysis (Appendix~\ref{sec:debate_rounds}) showing that both inter-agent agreement and accuracy largely plateau by Round 4, providing sufficient opportunity for answer revision while limiting additional computational cost.

Each debate team contains three agents: two weak agents and one strong agent. Three agents constitute the smallest team that supports majority voting while reducing the likelihood of ties. This is also a common team size in prior MAD work~\citep{zhang-etal-2024-exploring}. The two weak agents are separate instances of the same model, while the strong agent is an instance of a relatively more capable model. This creates a controlled mixed-capability setting in which a stronger minority interacts with a weaker majority. This configuration also reflects a practical use case in which several lighter-weight agents are paired with a stronger model for cost or efficiency reasons. When the strong and weak agents disagree, the final answer provides a natural way to assess the relative influence of the more capable minority and the less capable majority. We use two instances of the same weak model so that the weak side functions as a controlled unit, simplifying the experimental design. When varying speaking order, we change the strong agent's position relative to this fixed weak-agent block; when manipulating personality, both weak agents receive the same assignment. Our goal is to study how the strong side interacts with a controlled weak majority, rather than how differences among weak agents affect the debate. With heterogeneous weak agents, model identity, speaking position, and personality assignment would become additional factors that must be varied and disentangled, substantially increasing the number of experimental conditions and making the focal effects harder to isolate.

\subsection{Model Combinations}

To ensure that our findings hold across more than a specific model pair, we construct teams from seven commonly used LLMs spanning different sizes and generations across four model families: Google Gemini~\citep{comanici2025gemini25pushingfrontier}, OpenAI GPT~\citep{gpt4omini}, Meta Llama~\citep{meta2025llama4} and Mistral~\citep{mistral2025ministral14b}. From these models we form ten weak-strong combinations that include both within-family (e.g., Gemini 2.5 lite and Gemini 2.5) and cross-family pairings (e.g., GPT-4o-mini and Gemini 3). The weak and strong labels are assigned relative to each model pair, guided by expected model capability and empirically supported by standalone performance on the sampled benchmark questions: in every combination, the strong model achieves higher standalone accuracy than the weak model. In practice, relative capability can similarly be estimated from known differences in model scale and generation or standalone validation performance. Appendix~\ref{app:models_used} reports the list of models and combinations, together with their standalone accuracies. 

\subsection{Speaking Order and Personality Manipulation}
\label{sec:manipulation}

Our experimental design proceeds in two stages. We first isolate the effect of speaking order without personality prompting to identify whether sequential debate exhibits a positional bias. We then focus on the disadvantaged speaking order identified in this comparison and test whether personality-guided behavioral interventions can mitigate this disadvantage.

\textbf{Speaking order.} We evaluate two mirrored speaking orders: Weak-Weak-Strong (\textit{W-W-S}), where the strong agent speaks after the weak-agent block, and Strong-Weak-Weak (\textit{S-W-W}), where it speaks before the weak-agent block. This design keeps the two weak agents adjacent and preserves their relative interaction structure across conditions: the second weak agent always responds immediately after the first, ensuring that direct dependence between the weak agents is present in both orders. The primary change is therefore the strong agent’s position relative to this weak-agent block. Holding the model combination and task fixed, the comparison captures the total effect of this change in speaking order, including downstream changes in conversational content induced by the different order.

\textbf{Personality manipulation.} We study two Big Five dimensions separately: agreeableness and extraversion. All personality manipulations are conducted under the \textit{W-W-S} order, where the strong agent speaks last. For agents assigned a personality prompt, we set one target dimension to either high or low while holding the other four dimensions neutral, accompanied by behavioral instructions grounded in the corresponding trait (Appendix~\ref{app:prompts}). We treat these prompts as theory-guided behavioral configurations rather than as tests of stable human-like personality traits in LLMs. The personality prompt is applied either to the strong agent or to both weak agents; the other side receives no personality prompt. The two weak agents always receive the same personality assignment. This allows us to compare how changing the behavior of the strong side versus the weak side affects debate dynamics. Each dimension therefore yields four configurations: high or low trait levels applied to the strong or the weak side. To assess robustness to personality-prompt wording, we additionally test a simplified prompt without explicit trait-specific behavioral instructions using GPT-4o-mini/Gemini-3-Flash across all four benchmarks. Full details are provided in Appendix~\ref{app:alternative_personality_prompts}.

\textbf{Notation.} We denote each personality condition using a three-part shorthand that follows the agent sequence in the debate. The symbols \textit{ah} and \textit{al} denote high and low agreeableness, \textit{eh} and \textit{el} denote high and low extraversion, and \textit{n} denotes no personality prompt. For example, \textit{ah\_ah\_n} denotes the \textit{W-W-S} order where the two weak agents are set to high agreeableness and the strong agent receives no personality prompt. 

\subsection{Experimental Configurations}

A debate configuration is defined by a model combination, speaking order, and personality assignment. Each configuration is evaluated on all four benchmark datasets. Across configurations, we hold fixed the debate protocol, task prompts, and answer extraction methods. All questions are framed as zero-shot multiple-choice tasks, and all runs use temperature 0. Experiments are run through the OpenRouter API.\footnote{\url{https://openrouter.ai}} Our objective is to isolate the effects of speaking order and personality prompting rather than to outperform state-of-the-art MAD systems. If personality prompting proves to be an effective intervention, the mechanism can subsequently be integrated with existing MAD designs to further enhance performance. In addition to the debate configurations, we collect standalone single-agent responses from each model under every personality condition as accuracy baselines and pre-interaction reference answers.

\section{Metrics and Analytical Framework}
\label{sec:metrics}

For every question and debate condition, we record the answer choices produced by each agent in each round together with their textual justifications. We use these records to construct two primary outcome measures: final debate accuracy and agent influence.

\subsection{Metrics}

\textbf{Final debate accuracy} measures the collective reasoning performance of each debate condition. It is defined as the percentage of questions answered correctly according to the final round majority answer. If the final answers do not produce a majority consensus, the response is counted as incorrect. For single-agent settings, accuracy is defined analogously using the agent's standalone answer. Overall, 99.9\% of debates produced a final-round majority. 

\textbf{Agent influence} captures the extent to which each side shapes the final debate outcome when the strong and weak sides initially disagree. When the two sides begin from different answers and the debate ultimately adopts one of those answers, the side whose answer is adopted is treated as having prevailed in shaping the debate outcome. 

Formally, for each question let \(a_S\) denote the strong model's standalone answer, \(a_W\) the weak side's standalone answer, and \(a_F\) the final answer after debate. We compute influence over the set of questions where \(a_S \neq a_W\) and \(a_F \in \{a_S, a_W\}\). The first condition restricts attention to genuine disagreements: if the two sides start from the same answer, the final answer provides no basis for attributing the outcome to either side. The second condition excludes debates that converge on a third answer neither side held initially, since the outcome cannot be attributed cleanly to either side. Within this set, strong-agent influence is the percentage of questions for which \(a_F = a_S\), and weak-agent influence is the percentage for which \(a_F = a_W\). The two sum to 100\% by construction, making influence a direct, zero-sum measure of which side prevailed when the two sides initially disagree.

Using standalone answers as the reference point isolates each agent's pre-interaction position from responses produced within the debate, which may already be influenced by prior debate messages. Standalone responses are collected under the same personality condition as the corresponding debate. This baseline-versus-interaction comparison follows prior work on LLM attitude change~\citep{taubenfeld-etal-2024-systematic,ICLR2025_1da9ca7e}. Across the experimental data, around 45\% of questions fall into the subset used for the influence analysis, with the proportion remaining broadly stable across the different speaking-order and personality conditions.

\subsection{Analytical Strategy}
\label{sec:analytical_strategy}
We use matched comparisons to evaluate speaking-order and personality effects. Speaking-order effects are assessed by comparing the \textit{S-W-W} and \textit{W-W-S} conditions without personality prompting, matched by model combination and benchmark. Personality effects are assessed within the \textit{W-W-S} order by comparing each personality-prompted condition with the matched no-personality baseline, matched by model combination and benchmark. For each comparison, outcomes are aggregated within each model-combination--benchmark cell, yielding \(4\times10=40\) matched observations. We compute the paired percentage point difference in accuracy or influence for each observation and report the mean and standard deviation of these differences across the 40 cells. Statistical reliability is assessed using a paired \(t\)-test, a Wilcoxon signed-rank test, and a sign test. We also fit a mixed-effects model with experimental condition as a fixed effect and random intercepts for benchmark, model combination, and benchmark-by-model-combination cell. This provides an additional robustness check while accounting for heterogeneity across benchmarks, model combinations, and individual benchmark--model-combination cells. Influence tests are two-tailed because shifts in either direction are meaningful, whereas accuracy tests are one-tailed toward improvement. For the speaking-order comparison, the accuracy test evaluates whether \textit{S-W-W} outperforms \textit{W-W-S}; for the personality comparisons, it evaluates whether each personality-prompted \textit{W-W-S} condition improves over the matched no-personality \textit{W-W-S} baseline.

\section{Results}

\subsection{First-Speaker Bias}

\begin{figure}[h]
  \centering
  \includegraphics[width=0.48\linewidth]{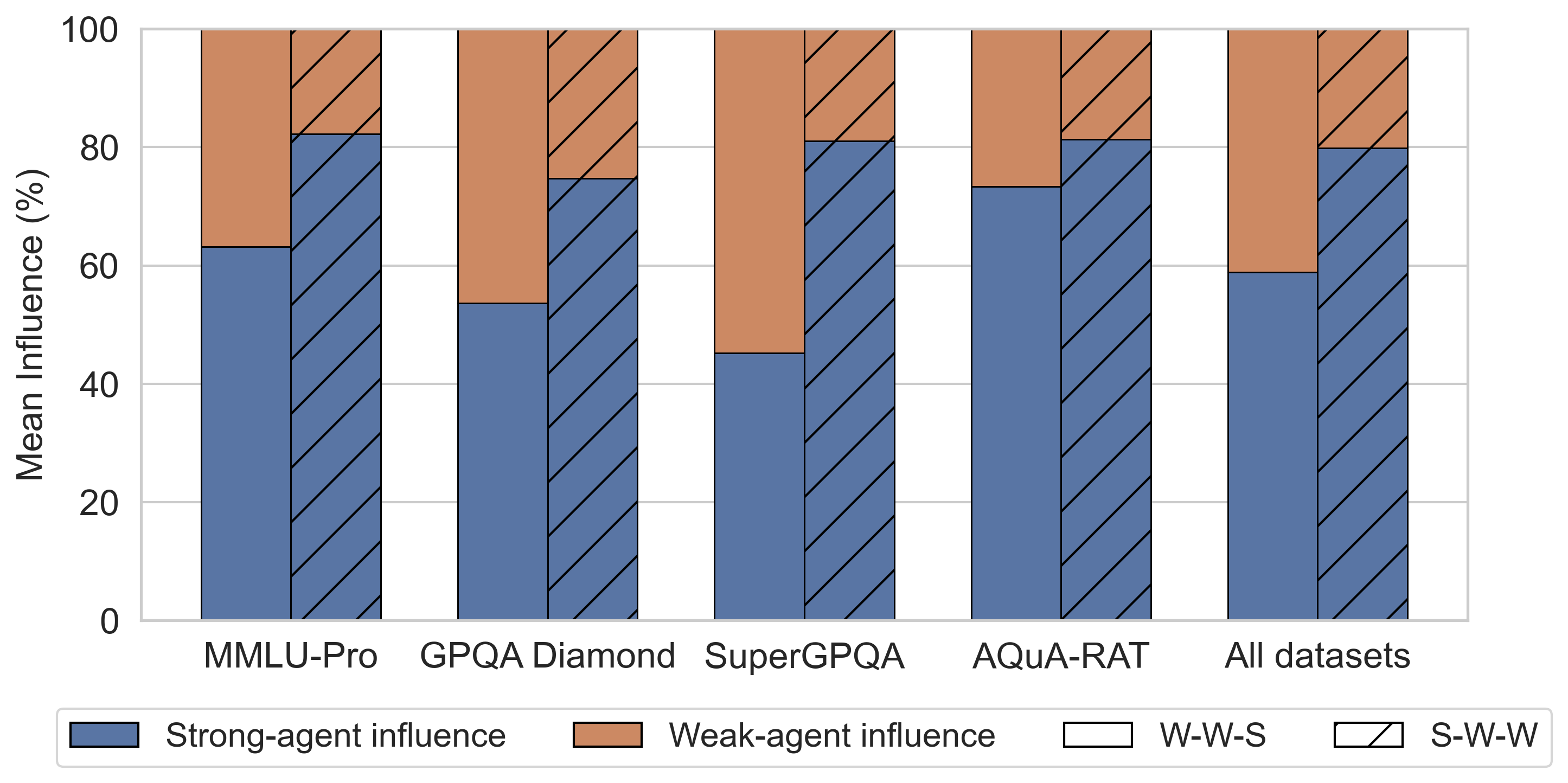}
  \hfill
  \includegraphics[width=0.48\linewidth]{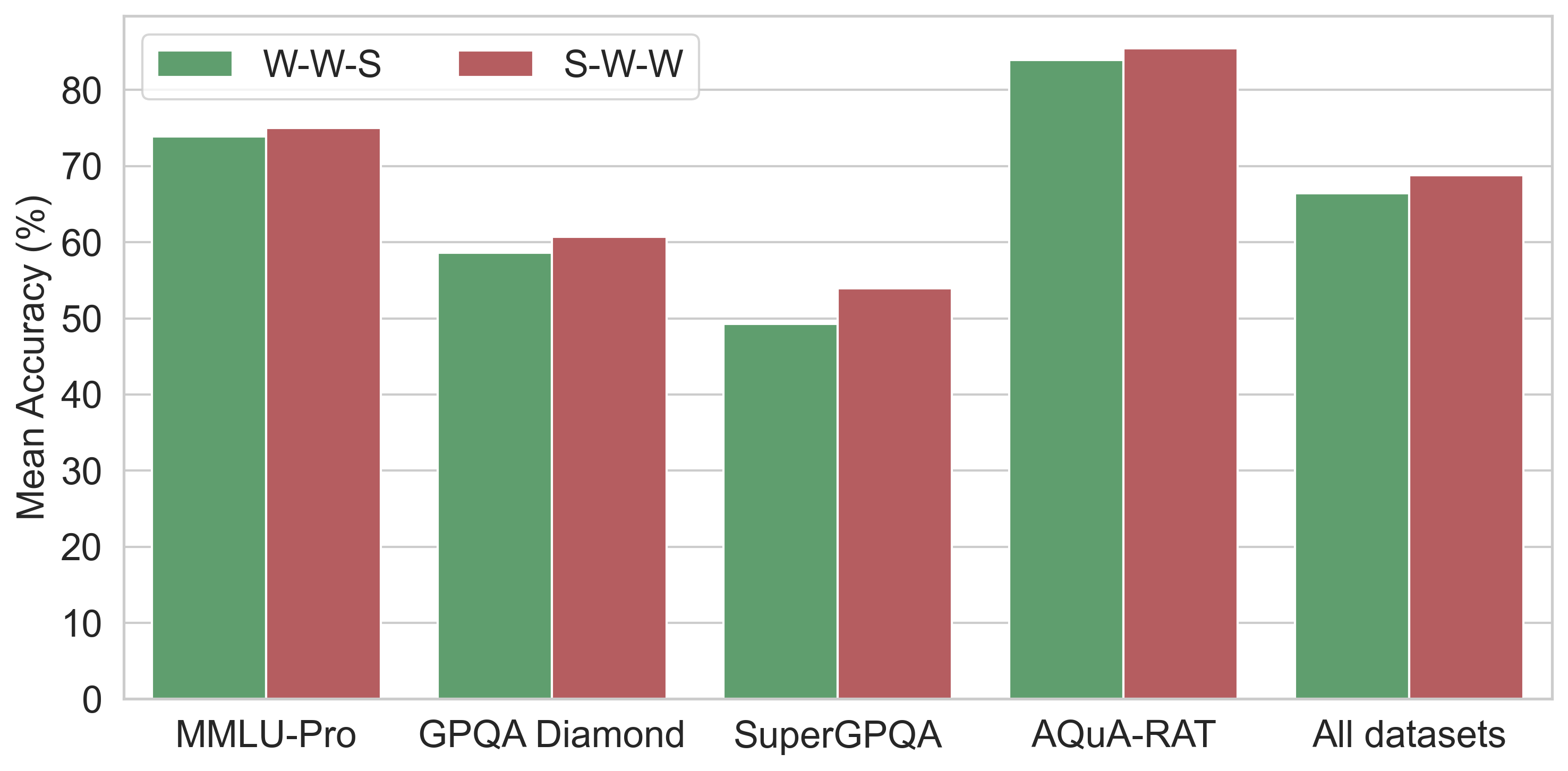}
  \caption{
    Results under the no-personality baseline, shown by benchmark dataset and averaged across model combinations.
    Left: Mean agent \textbf{influence} across debate orders, with stacked segments showing the relative influence of the strong and weak agents within each order.
    Right: Mean final debate \textbf{accuracy} across debate orders.
  }
  \label{fig:sequence_combined}
\end{figure}

We first compare the no-personality \textit{W-W-S} and \textit{S-W-W} baselines. Figure~\ref{fig:sequence_combined} shows that speaking order strongly affects agent influence. When the strong agent speaks first in \textit{S-W-W}, its influence is substantially higher than when it speaks last in \textit{W-W-S}. As reported in Table~\ref{tab:all_dataset_accuracy}, the strong-agent influence is 21.01 percentage points greater in \textit{S-W-W} than in \textit{W-W-S}. Speaking first therefore gives agents a systematic advantage in shaping the final majority answer.

\begin{table}[h]
  \caption{Average standalone and debate accuracy (\%), and strong-agent influence (\%) for the no-personality debate baselines, aggregated across benchmark
datasets and model combinations.}
  \label{tab:all_dataset_accuracy}
\centering
\begin{tabular}{lrrrr}
&\textbf{\textit{Weak}} & \textbf{\textit{Strong}} & \textbf{\textit{W-W-S}} & \textbf{\textit{S-W-W}} \\
\hline
Acc. & 41.29& 62.32& 66.43& 68.78\\
Infl. & N.A. & N.A. & 58.83& 79.84\\
\hline
\end{tabular}

\end{table}

This positional difference in influence is accompanied by a difference in reasoning performance. As shown in Figure~\ref{fig:sequence_combined}, \textit{S-W-W} achieves higher average accuracy than \textit{W-W-S} across all four benchmarks. Aggregated across datasets and model combinations, accuracy increases from 66.43\% under \textit{W-W-S} to 68.78\% under \textit{S-W-W}, a 2.34 percentage point difference as shown in Table~\ref{tab:all_dataset_accuracy}. The paired statistical tests and mixed-effects models further confirm that both final accuracy and strong-agent influence are significantly higher under \textit{S-W-W} than under \textit{W-W-S}; the full results are reported in Appendix~\ref{app:sequence_order_tests}.

We further examine whether this speaking-order effect varies with the capability gap between the strong and weak models. A mixed-effects moderation analysis shows that the influence disadvantage of placing the strong agent last decreases as the capability gap increases, whereas the accuracy difference between the two speaking orders does not significantly vary with the capability gap (Appendix~\ref{app:capability-gap}).

Taken together, these results identify a first-speaker bias: sequential debate is not a neutral aggregation procedure but a protocol whose influence structure is shaped by the order of interaction. In particular, placing the strong agent last reduces both its influence and debate accuracy relative to placing it first. We therefore focus the personality experiments on the disadvantaged \textit{W-W-S} configuration.

\subsection{Effects of Agreeableness}

As shown in Table~\ref{tab:agreeable_influence}, agreeableness produces a clear and systematic influence pattern: influence shifts in the direction of lower agreeableness. Since strong-agent and weak-agent influence sum to 100\%, changes in strong-agent influence also indicate corresponding shifts in weak-agent influence. When the strong agent is made less agreeable, its influence increases; when it is made more agreeable, its influence decreases. Conversely, making the weak agents more agreeable shifts influence toward the strong agent, while making them less agreeable shifts influence away from it. Most of these shifts are significant across the statistical tests, indicating that agreeableness provides a controllable behavioral lever for altering which side prevails when the strong and weak agents initially disagree.

\begin{table}[h]
\caption{Percentage point differences in strong-agent \textbf{influence} and final debate \textbf{accuracy} under \textbf{agreeableness} manipulations relative to the no-personality baseline in the Weak-Weak-Strong order, aggregated across benchmark datasets and model combinations. Accuracy tests are one-tailed, influence tests are two-tailed. Significance levels: $^{***}p<0.001$, $^{**}p<0.01$, $^{*}p<0.05$, $^{\dagger}p<0.1$.}
\label{tab:agreeable_influence}
\centering
\resizebox{\textwidth}{!}{%
\begin{tabular}{lrrrrrrrr}
& \multicolumn{4}{c}{\textbf{Strong-Agent Influence}} & \multicolumn{4}{c}{\textbf{Final Accuracy}} \\
\cline{2-5} \cline{6-9}
\textbf{Statistic}
& \textbf{\textit{ah\_ah\_n}}
& \textbf{\textit{al\_al\_n}}
& \textbf{\textit{n\_n\_ah}}
& \textbf{\textit{n\_n\_al}}
& \textbf{\textit{ah\_ah\_n}}
& \textbf{\textit{al\_al\_n}}
& \textbf{\textit{n\_n\_ah}}
& \textbf{\textit{n\_n\_al}} \\
\hline
Mean $\Delta$ (pp)
& 3.76& -4.84& -11.07& 9.55& -0.02& -2.42& -5.13& 1.64\\
Std
& 5.33& 9.77& 8.13& 6.89& 2.49& 4.11& 2.86& 3.42\\
Paired t-test $p$
& <0.001$^{***}$& 0.0033$^{**}$ & <0.001$^{***}$& <0.001$^{***}$
& 0.5231 & 0.9997 & 1.0000 & 0.0021$^{**}$\\
Wilcoxon $p$
& <0.001$^{***}$& 0.0120$^{*}$ & <0.001$^{***}$& <0.001$^{***}$
& 0.6171 & 0.9993 & 1.0000 & 0.0042$^{**}$\\
Sign test $p$
& 0.0022$^{**}$ & 0.2682 & <0.001$^{***}$& <0.001$^{***}$
& 0.3136 & 0.9996 & 1.0000 & 0.0069$^{**}$\\
Mixed effects $p$
& <0.001$^{***}$& 0.0017$^{**}$& <0.001$^{***}$& <0.001$^{***}$
& 0.5232& 0.9999& 1.0000 & 0.0012$^{**}$\\
\hline
\end{tabular}%
}
\end{table}

This redistribution of influence translates into accuracy improvements only under particular configurations. The largest improvement occurs when the strong agent is assigned low agreeableness while the weak agents remain unprompted (\textit{n\_n\_al}). Relative to the no-personality \textit{W-W-S} baseline, strong-agent influence rises by 9.55 percentage points and final accuracy by 1.64 percentage points, significant across all paired tests and the mixed-effects model. Thus, low agreeableness on the strong agent partially counteracts the positional disadvantage identified in RQ1, allowing the strong agent to retain greater influence despite speaking after the two weak agents. Appendix~\ref{app:debate_trace} presents a representative debate trace illustrating how greater resistance to preceding responses can produce this pattern. Conversely, high-agreeableness prompting on the strong agent (\textit{n\_n\_ah}) produces the largest reduction in both strong-agent influence and accuracy.

The relationship between influence and accuracy is expected given the capability structure of our teams. Because the strong model is more accurate than the weak model on average, shifting disagreements toward the strong model's answer should generally increase the likelihood of a correct outcome. Importantly, however, maximizing strong-agent influence is not itself the objective. The no-personality \textit{W-W-S} debate already outperforms the strong agent acting alone (Table~\ref{tab:all_dataset_accuracy}), suggesting that the weak agents can contribute useful information and gains can arise from the multi-agent interaction rather than simply deferring to the strongest model. We further test this interpretation using an additional configuration in which the strong agent is assigned low agreeableness while the weak agents are assigned high agreeableness (\textit{ah\_ah\_al}; Appendix~\ref{app:additional_agreeableness}). This configuration increases strong-agent influence beyond \textit{n\_n\_al} (71.43\% versus 68.38\%) but produces lower final accuracy (67.23\% versus 68.08\%). Similarly, \textit{ah\_ah\_n} significantly increases strong-agent influence without improving accuracy. Greater strong-agent influence therefore does not translate monotonically into better performance. Effective mitigation instead appears to require a balance: the strong agent must retain enough resistance to overcome its positional disadvantage without overly suppressing potentially useful contributions from the weak side.

Overall, agreeableness provides an effective intervention for the first-speaker bias identified in RQ1. Assigning low agreeableness to the strong agent while leaving the weak agents unprompted restores the strong agent's influence and improves final reasoning performance without eliminating the informational contribution of the weaker agents.

\subsection{Effects of Extraversion}

Compared with agreeableness, extraversion produces less systematic changes in influence. All four extraversion manipulations significantly shift strong-agent influence (Table~\ref{tab:extraversion_infl}), but the direction of these effects does not follow a consistent high-versus-low pattern. Neither high nor low extraversion therefore provides a predictable way to increase the prompted side's influence. Extraversion affects debate dynamics, but not through the systematic redistribution observed for agreeableness.

\begin{table}[h]
\caption{Percentage point differences in strong-agent \textbf{influence} and final debate \textbf{accuracy} under \textbf{extraversion} manipulations relative to the no-personality baseline in the Weak-Weak-Strong order, aggregated across benchmark datasets and model combinations. Accuracy tests are one-tailed, influence tests are two-tailed. Significance levels: $^{***}p<0.001$, $^{**}p<0.01$, $^{*}p<0.05$, $^{\dagger}p<0.1$.}
\label{tab:extraversion_infl}
\centering
\resizebox{\textwidth}{!}{%
\begin{tabular}{lrrrrrrrr}
& \multicolumn{4}{c}{\textbf{Strong-Agent Influence}} & \multicolumn{4}{c}{\textbf{Final Accuracy}} \\
\cline{2-5} \cline{6-9}
\textbf{Statistic}
& \textbf{\textit{eh\_eh\_n}}
& \textbf{\textit{el\_el\_n}}
& \textbf{\textit{n\_n\_eh}}
& \textbf{\textit{n\_n\_el}}
& \textbf{\textit{eh\_eh\_n}}
& \textbf{\textit{el\_el\_n}}
& \textbf{\textit{n\_n\_eh}}
& \textbf{\textit{n\_n\_el}} \\
\hline
Mean $\Delta$ (pp)
& 3.92& 3.94& -3.04& -4.14& 0.09& 0.78& -2.21& -2.53\\
Std
& 6.56& 5.09& 5.40& 6.32& 3.22& 2.67& 2.51& 2.31\\
Paired t-test $p$
& <0.001$^{***}$& <0.001$^{***}$& <0.001$^{***}$ & <0.001$^{***}$
& 0.4320 & 0.0365$^{*}$ & 1.0000 & 1.0000\\
Wilcoxon $p$
& <0.001$^{***}$ & <0.001$^{***}$& 0.0014$^{**}$ & <0.001$^{***}$
& 0.3500 & 0.0379$^{*}$ & 1.0000 & 1.0000\\
Sign test $p$
& 0.0022$^{**}$ & 0.0064$^{**}$ & 0.0166$^{*}$ & 0.0064$^{**}$
& 0.5660 & 0.0939$^{\dagger}$ & 1.0000 & 1.0000\\
Mixed effects $p$
& <0.001$^{***}$& <0.001$^{***}$& <0.001$^{***}$& <0.001$^{***}$
& 0.4316& 0.0327$^{*}$& 1.0000 & 1.0000\\
\hline
\end{tabular}%
}
\end{table}

\begin{figure}[h]
  \centering
  \includegraphics[width=0.65\linewidth]{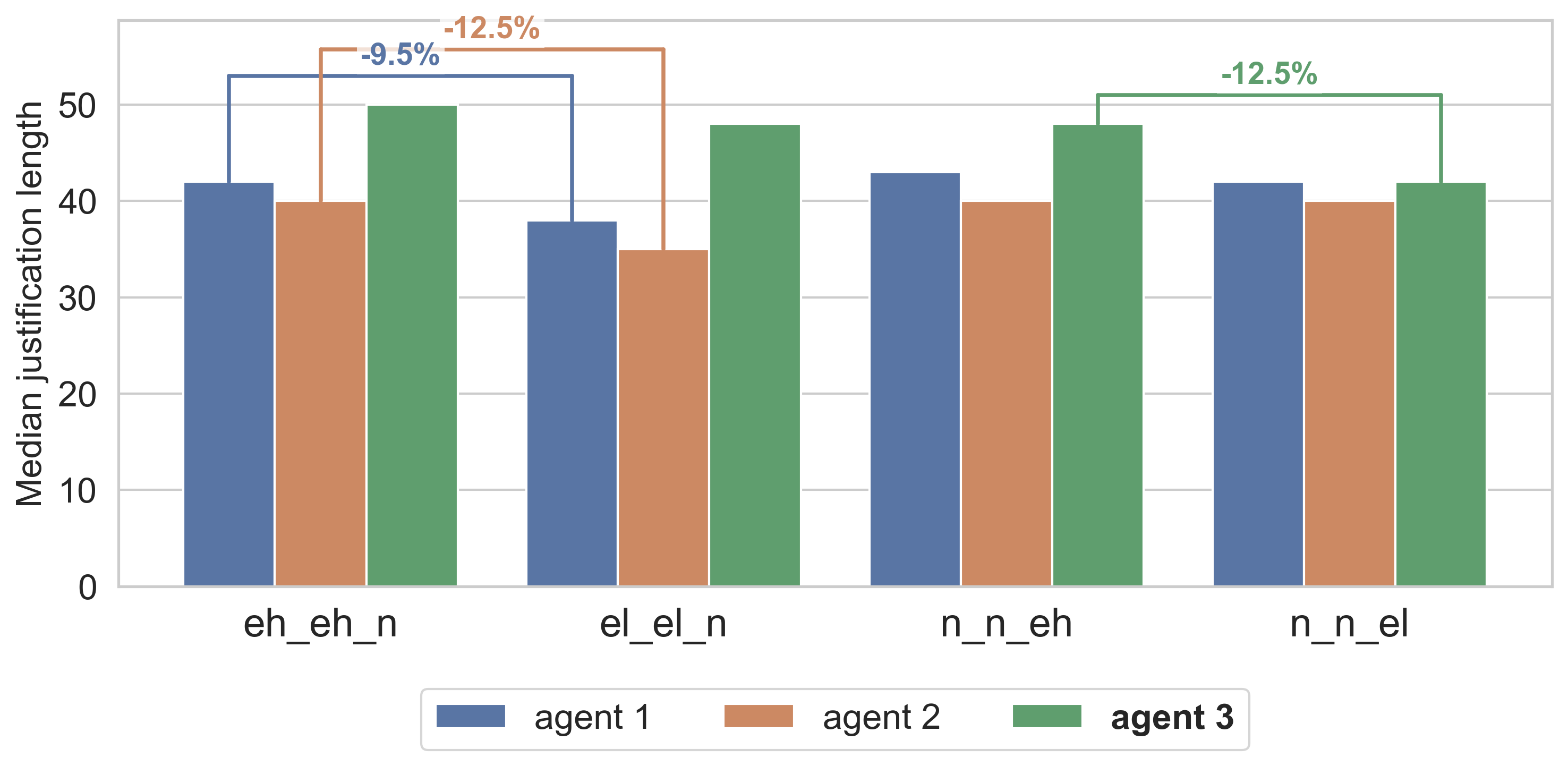}
  \caption{
    Median justification length by extraversion configuration and agent position in the \textit{W-W-S} order, aggregated across benchmark datasets and model combinations.
    Bold labels indicate the strong agent. Connected annotations show the percentage change in justification length between the high- and low-extraversion conditions, computed as $(\text{low} - \text{high}) / \text{high} \times 100\%$.
  }
  \label{fig:extraversion_just_combined}
\end{figure}

When the weak agents are assigned low extraversion, final accuracy improves by 0.78 percentage points. This improvement is significant under the paired $t$-test, Wilcoxon signed-rank test, and mixed-effects model, although the sign test provides only marginal evidence ($p=0.0939$). Compared with \textit{n\_n\_al} (+1.64 pp), this gain is smaller.

Because the effects of extraversion on influence and final accuracy are less systematic, we further examine whether the manipulation nevertheless produces observable changes in agents' debate behavior. Specifically, we analyze agents' verbosity using median justification length, since justification lengths are right-skewed. Figure~\ref{fig:extraversion_just_combined} shows that high-extraversion prompts generally lead agents to produce longer justifications than low-extraversion prompts in matched positions. This pattern appears across both strong- and weak-agent manipulations. This indicates that the manipulation is behaviorally meaningful: extraversion systematically changes how agents express their reasoning, even though these expressive differences do not translate into a consistent high-versus-low pattern in influence or final accuracy. By contrast, agreeableness does not systematically change justification length (Appendix~\ref{app:agree_just}), so its influence effects do not appear to be explained by a simple change in verbosity.

Taken together, the two personality dimensions produce qualitatively different effects. Agreeableness produces a systematic redistribution of influence in the direction of lower agreeableness. When low agreeableness is assigned to the stronger agent, this shift helps recover influence lost from speaking last and yields the largest accuracy improvement among the personality interventions. Extraversion produces more mixed changes in influence. Although one configuration also improves final accuracy, the gain is smaller and the statistical evidence is weaker than for the corresponding agreeableness improvement. Its clearest and most consistent behavioral effect is instead on how agents express their reasoning, with high-extraversion prompts generally producing longer justifications. The main directional pattern is also broadly reproduced under a simplified personality prompt without explicit trait-specific behavioral instructions (Appendix~\ref{app:alternative_personality_prompts}).

\section{Conclusion}

Sequential multi-agent debate is often used as if interaction simply aggregates the reasoning of multiple LLM agents---our results show that the protocol itself shapes the outcome. We identify a first-speaker bias: agents exert greater influence when they speak earlier, so placing the stronger agent after weaker agents reduces its influence and lowers final debate accuracy relative to placing it first. We then show that targeted personality-guided behavioral interventions can mitigate this positional disadvantage. Agreeableness provides a systematic lever for redistributing influence in the direction of lower agreeableness. In particular, assigning low agreeableness to the stronger agent while leaving the weaker agents unprompted restores some of its lost influence and yields the largest accuracy improvement among the personality interventions, while further increasing strong-agent influence does not necessarily produce additional performance gains. Extraversion, by contrast, produces less systematic changes in influence; although one configuration improves accuracy, its clearest and most consistent effect is on how agents express their reasoning. Effective MAD systems should therefore be designed as interaction protocols rather than merely ensembles of models: speaking order and personality-induced behavior jointly shape influence and collective reasoning outcomes.

\section*{Limitations}

This study has several limitations. First, we evaluate a controlled three-agent setting with two instances of a weak model and one instance of a strong model. This design supports clean comparisons of speaking order and influence, but does not capture larger teams or settings in which the weak side comprises distinct models.

Second, our model pool covers seven LLMs across four model families and ten weak--strong combinations. Although the combinations span different model families, generations, and capability gaps, broader coverage of model architectures and training paradigms would further strengthen the generalizability of the findings.

Third, our experiments focus on multiple-choice reasoning benchmarks. This setting enables clear measurement of final accuracy and answer-level influence, but may not capture interaction dynamics in more open-ended collaborative tasks such as writing, planning, or coding.

Finally, our personality manipulations should be interpreted as prompt-induced behavioral interventions rather than evidence of stable human-like personalities in LLMs. The main prompts contain explicit trait-specific behavioral instructions, so some effects may depend on prompt formulation or vary in magnitude across models. We therefore conduct an alternative-prompt robustness check without explicit behavioral instructions, which broadly reproduces the main directional pattern, although this check is limited to one model combination.

\subsection*{AI use statement}

We used generative AI tools to assist with code development and debugging and with editing parts of the manuscript for clarity and readability. All AI-assisted code was reviewed and tested by the authors, and all AI-assisted manuscript content was checked for accuracy and consistency with the underlying research and results. We have reviewed all AI-assisted work and take responsibility for the final content of this work, including text, claims, code, and other artifacts produced with the aid of generative AI.

\subsection*{Reproducibility statement}

We provide the information and materials needed to reproduce the experiments and analyses throughout the paper and appendices. Section~\ref{sec:data_sets} describes the benchmark datasets and question sampling, Sections~\ref{sec:design} and~\ref{sec:metrics} describe the debate protocol, experimental manipulations, evaluation metrics, and analytical strategy, and Appendix~\ref{app:models_used} reports the models, model combinations, and standalone performance used in the experiments. Appendix~\ref{app:prompts} provides the prompt templates, while the remaining appendices report additional statistical analyses, robustness checks, and per-dataset results. The anonymous repository linked in the abstract contains the code used to run the experiments and reproduce the reported analyses.

\bibliography{iclr2027_conference}

@inproceedings{rein2024gpqa,
title={{GPQA}: A Graduate-Level Google-Proof Q\&A Benchmark},
author={David Rein and Betty Li Hou and Asa Cooper Stickland and Jackson Petty and Richard Yuanzhe Pang and Julien Dirani and Julian Michael and Samuel R. Bowman},
booktitle={First Conference on Language Modeling},
year={2024},
url={https://openreview.net/forum?id=Ti67584b98}
}

@article{barryBargainerCharacteristicsDistributive1998a,
  title = {Bargainer Characteristics in Distributive and Integrative Negotiation.},
  author = {Barry, Bruce and Friedman, Raymond A.},
  year = 1998,
  journal = {Journal of Personality and Social Psychology},
  volume = {74},
  number = {2},
  pages = {345--359},
  publisher = {American Psychological Association},
  address = {US},
  issn = {1939-1315(Electronic),0022-3514(Print)},
  doi = {10.1037/0022-3514.74.2.345}
}

@inproceedings{ICLR2024_578e65cd,
 author = {Chen, Weize and Su, Yusheng and Zuo, Jingwei and Yang, Cheng and Yuan, Chenfei and Chan, Chi-Min and Yu, Heyang and Lu, Yaxi and Hung, Yi-Hsin and Qian, Chen and Qin, Yujia and Cong, Xin and Xie, Ruobing and Liu, Zhiyuan and Sun, Maosong and Zhou, Jie},
 booktitle = {International Conference on Learning Representations},
 editor = {B. Kim and Y. Yue and S. Chaudhuri and K. Fragkiadaki and M. Khan and Y. Sun},
 pages = {20094--20136},
 title = {AgentVerse: Facilitating Multi-Agent Collaboration and Exploring Emergent Behaviors},
 url = {https://proceedings.iclr.cc/paper_files/paper/2024/file/578e65cdee35d00c708d4c64bce32971-Paper-Conference.pdf},
 volume = {2024},
 year = {2024}
}

@inproceedings{ICLR2024_25cc3adf,
 author = {Chan, Chi-Min and Chen, Weize and Su, Yusheng and Yu, Jianxuan and Xue, Wei and Zhang, Shanghang and Fu, Jie and Liu, Zhiyuan},
 booktitle = {International Conference on Learning Representations},
 editor = {B. Kim and Y. Yue and S. Chaudhuri and K. Fragkiadaki and M. Khan and Y. Sun},
 pages = {9079--9093},
 title = {ChatEval: Towards Better LLM-based Evaluators through Multi-Agent Debate},
 url = {https://proceedings.iclr.cc/paper_files/paper/2024/file/25cc3adf8c85f7c70989cb8a97a691a7-Paper-Conference.pdf},
 volume = {2024},
 year = {2024}
}

@misc{cho2025herdbehaviorinvestigatingpeer,
      title={Herd Behavior: Investigating Peer Influence in LLM-based Multi-Agent Systems}, 
      author={Young-Min Cho and Sharath Chandra Guntuku and Lyle Ungar},
      year={2025},
      eprint={2505.21588},
      archivePrefix={arXiv},
      primaryClass={cs.MA},
      url={https://arxiv.org/abs/2505.21588}, 
}

@inproceedings{choi-etal-2025-empirical,
    title = "An Empirical Study of Group Conformity in Multi-Agent Systems",
    author = "Choi, Min  and
      Kim, Keonwoo  and
      Chae, Sungwon  and
      Baek, Sangyeop",
    editor = "Che, Wanxiang  and
      Nabende, Joyce  and
      Shutova, Ekaterina  and
      Pilehvar, Mohammad Taher",
    booktitle = "Findings of the Association for Computational Linguistics: ACL 2025",
    month = jul,
    year = "2025",
    address = "Vienna, Austria",
    publisher = "Association for Computational Linguistics",
    url = "https://aclanthology.org/2025.findings-acl.265/",
    doi = "10.18653/v1/2025.findings-acl.265",
    pages = "5123--5139",
    ISBN = "979-8-89176-256-5"
}

@inproceedings{cisneros-velarde-2025-biases,
    title = "Biases in Opinion Dynamics in Multi-Agent Systems of Large Language Models: A Case Study on Funding Allocation",
    author = "Cisneros-Velarde, Pedro",
    editor = "Chiruzzo, Luis  and
      Ritter, Alan  and
      Wang, Lu",
    booktitle = "Findings of the Association for Computational Linguistics: NAACL 2025",
    month = apr,
    year = "2025",
    address = "Albuquerque, New Mexico",
    publisher = "Association for Computational Linguistics",
    url = "https://aclanthology.org/2025.findings-naacl.101/",
    doi = "10.18653/v1/2025.findings-naacl.101",
    pages = "1889--1916",
    ISBN = "979-8-89176-195-7"
}

@article{costaDomainsFacetsHierarchical1995,
  title = {Domains and Facets: {{Hierarchical}} Personality Assessment Using the {{Revised NEO Personality Inventory}}.},
  author = {Costa, Paul T. and McCrae, Robert R.},
  year = 1995,
  journal = {Journal of Personality Assessment},
  volume = {64},
  number = {1},
  pages = {21--50},
  publisher = {Lawrence Erlbaum},
  address = {US},
  issn = {1532-7752(Electronic),0022-3891(Print)},
  doi = {10.1207/s15327752jpa6401_2}
}

@InProceedings{pmlr-v235-du24e,
  title = 	 {Improving Factuality and Reasoning in Language Models through Multiagent Debate},
  author =       {Du, Yilun and Li, Shuang and Torralba, Antonio and Tenenbaum, Joshua B. and Mordatch, Igor},
  booktitle = 	 {Proceedings of the 41st International Conference on Machine Learning},
  pages = 	 {11733--11763},
  year = 	 {2024},
  editor = 	 {Salakhutdinov, Ruslan and Kolter, Zico and Heller, Katherine and Weller, Adrian and Oliver, Nuria and Scarlett, Jonathan and Berkenkamp, Felix},
  volume = 	 {235},
  series = 	 {Proceedings of Machine Learning Research},
  month = 	 {21--27 Jul},
  publisher =    {PMLR},
  url = 	 {https://proceedings.mlr.press/v235/du24e.html}
}

@inproceedings{lin-hooi-2025-enhancing,
    title = "Enhancing Multi-Agent Debate System Performance via Confidence Expression",
    author = "Lin, Zijie  and
      Hooi, Bryan",
    editor = "Christodoulopoulos, Christos  and
      Chakraborty, Tanmoy  and
      Rose, Carolyn  and
      Peng, Violet",
    booktitle = "Findings of the Association for Computational Linguistics: EMNLP 2025",
    month = nov,
    year = "2025",
    address = "Suzhou, China",
    publisher = "Association for Computational Linguistics",
    url = "https://aclanthology.org/2025.findings-emnlp.343/",
    doi = "10.18653/v1/2025.findings-emnlp.343",
    pages = "6453--6471",
    ISBN = "979-8-89176-335-7"
}

@inproceedings{echterhoff-etal-2024-cognitive,
    title = "Cognitive Bias in Decision-Making with {LLM}s",
    author = "Echterhoff, Jessica Maria  and
      Liu, Yao  and
      Alessa, Abeer  and
      McAuley, Julian  and
      He, Zexue",
    editor = "Al-Onaizan, Yaser  and
      Bansal, Mohit  and
      Chen, Yun-Nung",
    booktitle = "Findings of the Association for Computational Linguistics: EMNLP 2024",
    month = nov,
    year = "2024",
    address = "Miami, Florida, USA",
    publisher = "Association for Computational Linguistics",
    url = "https://aclanthology.org/2024.findings-emnlp.739/",
    doi = "10.18653/v1/2024.findings-emnlp.739",
    pages = "12640--12653"
}

@inproceedings{NEURIPS2024_32e07a11,
 author = {Estornell, Andrew and Liu, Yang},
 booktitle = {Advances in Neural Information Processing Systems},
 doi = {10.52202/079017-0911},
 editor = {A. Globerson and L. Mackey and D. Belgrave and A. Fan and U. Paquet and J. Tomczak and C. Zhang},
 pages = {28938--28964},
 publisher = {Curran Associates, Inc.},
 title = {Multi-LLM Debate: Framework, Principals, and Interventions},
 url = {https://proceedings.neurips.cc/paper_files/paper/2024/file/32e07a110c6c6acf1afbf2bf82b614ad-Paper-Conference.pdf},
 volume = {37},
 year = {2024}
}

@inproceedings{fang-etal-2025-counterfactual,
    title = "Counterfactual Debating with Preset Stances for Hallucination Elimination of {LLM}s",
    author = "Fang, Yi  and
      Li, Moxin  and
      Wang, Wenjie  and
      Hui, Lin  and
      Feng, Fuli",
    editor = "Rambow, Owen  and
      Wanner, Leo  and
      Apidianaki, Marianna  and
      Al-Khalifa, Hend  and
      Eugenio, Barbara Di  and
      Schockaert, Steven",
    booktitle = "Proceedings of the 31st International Conference on Computational Linguistics",
    month = jan,
    year = "2025",
    address = "Abu Dhabi, UAE",
    publisher = "Association for Computational Linguistics",
    url = "https://aclanthology.org/2025.coling-main.703/",
    pages = "10554--10568"
}

@inproceedings{frisch-giulianelli-2024-llm,
    title = "{LLM} Agents in Interaction: Measuring Personality Consistency and Linguistic Alignment in Interacting Populations of Large Language Models",
    author = "Frisch, Ivar  and
      Giulianelli, Mario",
    editor = "Deshpande, Ameet  and
      Hwang, EunJeong  and
      Murahari, Vishvak  and
      Park, Joon Sung  and
      Yang, Diyi  and
      Sabharwal, Ashish  and
      Narasimhan, Karthik  and
      Kalyan, Ashwin",
    booktitle = "Proceedings of the 1st Workshop on Personalization of Generative AI Systems (PERSONALIZE 2024)",
    month = mar,
    year = "2024",
    address = "St. Julians, Malta",
    publisher = "Association for Computational Linguistics",
    url = "https://aclanthology.org/2024.personalize-1.9/",
    doi = "10.18653/v1/2024.personalize-1.9",
    pages = "102--111"
}

@article{goldbergDevelopmentMarkersBigFive1992a,
  title = {The Development of Markers for the {{Big-Five}} Factor Structure.},
  author = {Goldberg, Lewis R.},
  year = 1992,
  journal = {Psychological Assessment},
  volume = {4},
  number = {1},
  pages = {26--42},
  publisher = {American Psychological Association},
  address = {US},
  issn = {1939-134X(Electronic),1040-3590(Print)},
  doi = {10.1037/1040-3590.4.1.26}
}

@inproceedings{gu-etal-2025-explain,
    title = "Explain-Analyze-Generate: A Sequential Multi-Agent Collaboration Method for Complex Reasoning",
    author = "Gu, WenYuan  and
      Han, JiaLe  and
      Wang, HaoWen  and
      Li, Xiang  and
      Cheng, Bo",
    editor = "Rambow, Owen  and
      Wanner, Leo  and
      Apidianaki, Marianna  and
      Al-Khalifa, Hend  and
      Eugenio, Barbara Di  and
      Schockaert, Steven",
    booktitle = "Proceedings of the 31st International Conference on Computational Linguistics",
    month = jan,
    year = "2025",
    address = "Abu Dhabi, UAE",
    publisher = "Association for Computational Linguistics",
    url = "https://aclanthology.org/2025.coling-main.475/",
    pages = "7127--7140"
}

@inproceedings{ICLR2024_6507b115,
 author = {Hong, Sirui and Zhuge, Mingchen and Chen, Jonathan and Zheng, Xiawu and Cheng, Yuheng and Wang, Jinlin and Zhang, Ceyao and Wang, Zili and Yau, Steven and Lin, Zijuan and Zhou, Liyang and Ran, Chenyu and Xiao, Lingfeng and Wu, Chenglin and Schmidhuber, J\"{u}rgen},
 booktitle = {International Conference on Learning Representations},
 editor = {B. Kim and Y. Yue and S. Chaudhuri and K. Fragkiadaki and M. Khan and Y. Sun},
 pages = {23247--23275},
 title = {MetaGPT: Meta Programming for A Multi-Agent Collaborative Framework},
 url = {https://proceedings.iclr.cc/paper_files/paper/2024/file/6507b115562bb0a305f1958ccc87355a-Paper-Conference.pdf},
 volume = {2024},
 year = {2024}
}

@inproceedings{huang-hadfi-2024-personality,
    title = "How Personality Traits Influence Negotiation Outcomes? A Simulation based on Large Language Models",
    author = "Huang, Yin Jou  and
      Hadfi, Rafik",
    editor = "Al-Onaizan, Yaser  and
      Bansal, Mohit  and
      Chen, Yun-Nung",
    booktitle = "Findings of the Association for Computational Linguistics: EMNLP 2024",
    month = nov,
    year = "2024",
    address = "Miami, Florida, USA",
    publisher = "Association for Computational Linguistics",
    url = "https://aclanthology.org/2024.findings-emnlp.605/",
    doi = "10.18653/v1/2024.findings-emnlp.605",
    pages = "10336--10351"
}

@inproceedings{NEURIPS2023_21f7b745,
 author = {Jiang, Guangyuan and Xu, Manjie and Zhu, Song-Chun and Han, Wenjuan and Zhang, Chi and Zhu, Yixin},
 booktitle = {Advances in Neural Information Processing Systems},
 editor = {A. Oh and T. Naumann and A. Globerson and K. Saenko and M. Hardt and S. Levine},
 pages = {10622--10643},
 publisher = {Curran Associates, Inc.},
 title = {Evaluating and Inducing Personality in Pre-trained Language Models},
 url = {https://proceedings.neurips.cc/paper_files/paper/2023/file/21f7b745f73ce0d1f9bcea7f40b1388e-Paper-Conference.pdf},
 volume = {36},
 year = {2023}
}

@inproceedings{jiang-etal-2024-personallm,
    title = "{P}ersona{LLM}: Investigating the Ability of Large Language Models to Express Personality Traits",
    author = "Jiang, Hang  and
      Zhang, Xiajie  and
      Cao, Xubo  and
      Breazeal, Cynthia  and
      Roy, Deb  and
      Kabbara, Jad",
    editor = "Duh, Kevin  and
      Gomez, Helena  and
      Bethard, Steven",
    booktitle = "Findings of the Association for Computational Linguistics: NAACL 2024",
    month = jun,
    year = "2024",
    address = "Mexico City, Mexico",
    publisher = "Association for Computational Linguistics",
    url = "https://aclanthology.org/2024.findings-naacl.229/",
    doi = "10.18653/v1/2024.findings-naacl.229",
    pages = "3605--3627"
}

@inproceedings{kaesberg-etal-2025-voting,
    title = "Voting or Consensus? Decision-Making in Multi-Agent Debate",
    author = "Kaesberg, Lars Benedikt  and
      Becker, Jonas  and
      Wahle, Jan Philip  and
      Ruas, Terry  and
      Gipp, Bela",
    editor = "Che, Wanxiang  and
      Nabende, Joyce  and
      Shutova, Ekaterina  and
      Pilehvar, Mohammad Taher",
    booktitle = "Findings of the Association for Computational Linguistics: ACL 2025",
    month = jul,
    year = "2025",
    address = "Vienna, Austria",
    publisher = "Association for Computational Linguistics",
    url = "https://aclanthology.org/2025.findings-acl.606/",
    doi = "10.18653/v1/2025.findings-acl.606",
    pages = "11640--11671",
    ISBN = "979-8-89176-256-5"
}

@InProceedings{pmlr-v235-khan24a,
  title = 	 {Debating with More Persuasive {LLM}s Leads to More Truthful Answers},
  author =       {Khan, Akbir and Hughes, John and Valentine, Dan and Ruis, Laura and Sachan, Kshitij and Radhakrishnan, Ansh and Grefenstette, Edward and Bowman, Samuel R. and Rockt\"{a}schel, Tim and Perez, Ethan},
  booktitle = 	 {Proceedings of the 41st International Conference on Machine Learning},
  pages = 	 {23662--23733},
  year = 	 {2024},
  editor = 	 {Salakhutdinov, Ruslan and Kolter, Zico and Heller, Katherine and Weller, Adrian and Oliver, Nuria and Scarlett, Jonathan and Berkenkamp, Felix},
  volume = 	 {235},
  series = 	 {Proceedings of Machine Learning Research},
  month = 	 {21--27 Jul},
  publisher =    {PMLR},
  url = 	 {https://proceedings.mlr.press/v235/khan24a.html}
}

@inproceedings{NEURIPS2023_a3621ee9,
 author = {Li, Guohao and Hammoud, Hasan and Itani, Hani and Khizbullin, Dmitrii and Ghanem, Bernard},
 booktitle = {Advances in Neural Information Processing Systems},
 editor = {A. Oh and T. Naumann and A. Globerson and K. Saenko and M. Hardt and S. Levine},
 pages = {51991--52008},
 publisher = {Curran Associates, Inc.},
 title = {CAMEL: Communicative Agents for "Mind" Exploration of Large Language Model Society},
 url = {https://proceedings.neurips.cc/paper_files/paper/2023/file/a3621ee907def47c1b952ade25c67698-Paper-Conference.pdf},
 volume = {36},
 year = {2023}
}

@inproceedings{liang-etal-2024-encouraging,
    title = "Encouraging Divergent Thinking in Large Language Models through Multi-Agent Debate",
    author = "Liang, Tian  and
      He, Zhiwei  and
      Jiao, Wenxiang  and
      Wang, Xing  and
      Wang, Yan  and
      Wang, Rui  and
      Yang, Yujiu  and
      Shi, Shuming  and
      Tu, Zhaopeng",
    editor = "Al-Onaizan, Yaser  and
      Bansal, Mohit  and
      Chen, Yun-Nung",
    booktitle = "Proceedings of the 2024 Conference on Empirical Methods in Natural Language Processing",
    month = nov,
    year = "2024",
    address = "Miami, Florida, USA",
    publisher = "Association for Computational Linguistics",
    url = "https://aclanthology.org/2024.emnlp-main.992/",
    doi = "10.18653/v1/2024.emnlp-main.992",
    pages = "17889--17904"
}

@inproceedings{
liu2024a,
title={A Dynamic {LLM}-Powered Agent Network for Task-Oriented Agent Collaboration},
author={Zijun Liu and Yanzhe Zhang and Peng Li and Yang Liu and Diyi Yang},
booktitle={First Conference on Language Modeling},
year={2024},
url={https://openreview.net/forum?id=XII0Wp1XA9}
}

@article{Ma_Gao_Chai_Sun_Wang_Pei_Tao_Song_Liu_Zhang_Cui_2025, 
title={Debate on Graph: A Flexible and Reliable Reasoning Framework for Large Language Models}, 
volume={39}, 
url={https://ojs.aaai.org/index.php/AAAI/article/view/34658}, DOI={10.1609/aaai.v39i23.34658}, 
abstractNote={Large Language Models (LLMs) may suffer from hallucinations in real-world applications due to the lack of relevant knowledge. In contrast, knowledge graphs encompass extensive, multi-relational structures that store a vast array of symbolic facts. Consequently, integrating LLMs with knowledge graphs has been extensively explored, with Knowledge Graph Question Answering (KGQA) serving as a critical touchstone for the integration. This task requires LLMs to answer natural language questions by retrieving relevant triples from knowledge graphs. However, existing methods face two significant challenges: *excessively long reasoning paths distracting from the answer generation*, and *false-positive relations hindering the path refinement*. In this paper, we propose an iterative interactive KGQA framework that leverages the interactive learning capabilities of LLMs to perform reasoning and Debating over Graphs (DoG). Specifically, DoG employs a subgraph-focusing mechanism, allowing LLMs to perform answer trying after each reasoning step, thereby mitigating the impact of lengthy reasoning paths. On the other hand, DoG utilizes a multi-role debate team to gradually simplify complex questions, reducing the influence of false-positive relations. This debate mechanism ensures the reliability of the reasoning process. Experimental results on five public datasets demonstrate the effectiveness and superiority of our architecture. Notably, DoG outperforms the state-of-the-art method ToG by 23.7% and 9.1% in accuracy on WebQuestions and GrailQA, respectively. Furthermore, the integration experiments with various LLMs on the mentioned datasets highlight the flexibility of DoG.}, number={23}, journal={Proceedings of the AAAI Conference on Artificial Intelligence}, author={Ma, Jie and Gao, Zhitao and Chai, Qi and Sun, Wangchun and Wang, Pinghui and Pei, Hongbin and Tao, Jing and Song, Lingyun and Liu, Jun and Zhang, Chen and Cui, Lizhen}, year={2025}, month={Apr.}, pages={24768-24776} }

@inproceedings{
oh2025when,
title={When Debate Fails: Bias Reinforcement in Large Language Models},
author={Jihwan Oh and Minchan Jeong and Jongwoo Ko and Se-Young Yun},
booktitle={Workshop on Reasoning and Planning for Large Language Models},
year={2025},
url={https://openreview.net/forum?id=c5bjw7hqix}
}

@inproceedings{pitre-etal-2025-consensagent,
    title = "{CONSENSAGENT}: Towards Efficient and Effective Consensus in Multi-Agent {LLM} Interactions Through Sycophancy Mitigation",
    author = "Pitre, Priya  and
      Ramakrishnan, Naren  and
      Wang, Xuan",
    editor = "Che, Wanxiang  and
      Nabende, Joyce  and
      Shutova, Ekaterina  and
      Pilehvar, Mohammad Taher",
    booktitle = "Findings of the Association for Computational Linguistics: ACL 2025",
    month = jul,
    year = "2025",
    address = "Vienna, Austria",
    publisher = "Association for Computational Linguistics",
    url = "https://aclanthology.org/2025.findings-acl.1141/",
    doi = "10.18653/v1/2025.findings-acl.1141",
    pages = "22112--22133",
    ISBN = "979-8-89176-256-5"
}

@misc{serapiogarcia2025personalitytraitslargelanguage,
      title={Personality Traits in Large Language Models}, 
      author={Greg Serapio-García and Mustafa Safdari and Clément Crepy and Luning Sun and Stephen Fitz and Peter Romero and Marwa Abdulhai and Aleksandra Faust and Maja Matarić},
      year={2025},
      eprint={2307.00184},
      archivePrefix={arXiv},
      primaryClass={cs.CL},
      url={https://arxiv.org/abs/2307.00184}, 
}

@inproceedings{NEURIPS2025_a3c5af1f,
 author = {Du, Xeron and Yao, Yifan and Ma, Kaijing and Wang, Bingli and Zheng, Tianyu and Zhu, King and Liu, Minghao and others},
 booktitle = {Advances in Neural Information Processing Systems},
 editor = {D. Belgrave and C. Zhang and H. Lin and R. Pascanu and P. Koniusz and M. Ghassemi and N. Chen},
 pages = {},
 publisher = {Curran Associates, Inc.},
 title = {SuperGPQA: Scaling LLM Evaluation across 285 Graduate Disciplines},
 url = {https://proceedings.neurips.cc/paper_files/paper/2025/file/a3c5af1f56fc73eef1ba0f442739f5ca-Paper-Datasets_and_Benchmarks_Track.pdf},
 volume = {38},
 year = {2025}
}

@inproceedings{NEURIPS2024_ad236edc,
 author = {Wang, Yubo and Ma, Xueguang and Zhang, Ge and Ni, Yuansheng and Chandra, Abhranil and Guo, Shiguang and Ren, Weiming and Arulraj, Aaran and He, Xuan and Jiang, Ziyan and Li, Tianle and Ku, Max and Wang, Kai and Zhuang, Alex and Fan, Rongqi and Yue, Xiang and Chen, Wenhu},
 booktitle = {Advances in Neural Information Processing Systems},
 doi = {10.52202/079017-3018},
 editor = {A. Globerson and L. Mackey and D. Belgrave and A. Fan and U. Paquet and J. Tomczak and C. Zhang},
 pages = {95266--95290},
 publisher = {Curran Associates, Inc.},
 title = {MMLU-Pro: A More Robust and Challenging Multi-Task Language Understanding Benchmark},
 url = {https://proceedings.neurips.cc/paper_files/paper/2024/file/ad236edc564f3e3156e1b2feafb99a24-Paper-Datasets_and_Benchmarks_Track.pdf},
 volume = {37},
 year = {2024}
}

@inproceedings{ICLR2025_1da9ca7e,
 author = {Weng, Zhiyuan and Chen, Guikun and Wang, Wenguan},
 booktitle = {International Conference on Learning Representations},
 editor = {Y. Yue and A. Garg and N. Peng and F. Sha and R. Yu},
 pages = {11022--11060},
 title = {Do as We Do, Not as You Think: the Conformity of Large Language Models},
 url = {https://proceedings.iclr.cc/paper_files/paper/2025/file/1da9ca7e9cef4b1af63913f05d1630a4-Paper-Conference.pdf},
 volume = {2025},
 year = {2025}
}

@inproceedings{zhang-xiong-2025-debate4math,
    title = "{D}ebate4{MATH}: Multi-Agent Debate for Fine-Grained Reasoning in Math",
    author = "Zhang, Shaowei  and
      Xiong, Deyi",
    editor = "Che, Wanxiang  and
      Nabende, Joyce  and
      Shutova, Ekaterina  and
      Pilehvar, Mohammad Taher",
    booktitle = "Findings of the Association for Computational Linguistics: ACL 2025",
    month = jul,
    year = "2025",
    address = "Vienna, Austria",
    publisher = "Association for Computational Linguistics",
    url = "https://aclanthology.org/2025.findings-acl.862/",
    doi = "10.18653/v1/2025.findings-acl.862",
    pages = "16810--16824",
    ISBN = "979-8-89176-256-5"
}

@inproceedings{zhong-etal-2024-agieval,
    title = "{AGIE}val: A Human-Centric Benchmark for Evaluating Foundation Models",
    author = "Zhong, Wanjun  and
      Cui, Ruixiang  and
      Guo, Yiduo  and
      Liang, Yaobo  and
      Lu, Shuai  and
      Wang, Yanlin  and
      Saied, Amin  and
      Chen, Weizhu  and
      Duan, Nan",
    editor = "Duh, Kevin  and
      Gomez, Helena  and
      Bethard, Steven",
    booktitle = "Findings of the Association for Computational Linguistics: NAACL 2024",
    month = jun,
    year = "2024",
    address = "Mexico City, Mexico",
    publisher = "Association for Computational Linguistics",
    url = "https://aclanthology.org/2024.findings-naacl.149/",
    doi = "10.18653/v1/2024.findings-naacl.149",
    pages = "2299--2314"
}

@inproceedings{zhu-etal-2025-conformity,
    title = "Conformity in Large Language Models",
    author = "Zhu, Xiaochen  and
      Zhang, Caiqi  and
      Stafford, Tom  and
      Collier, Nigel  and
      Vlachos, Andreas",
    editor = "Che, Wanxiang  and
      Nabende, Joyce  and
      Shutova, Ekaterina  and
      Pilehvar, Mohammad Taher",
    booktitle = "Proceedings of the 63rd Annual Meeting of the Association for Computational Linguistics (Volume 1: Long Papers)",
    month = jul,
    year = "2025",
    address = "Vienna, Austria",
    publisher = "Association for Computational Linguistics",
    url = "https://aclanthology.org/2025.acl-long.195/",
    doi = "10.18653/v1/2025.acl-long.195",
    pages = "3854--3872",
    ISBN = "979-8-89176-251-0"
}

@inproceedings{zhang-etal-2024-exploring,
    title = "Exploring Collaboration Mechanisms for {LLM} Agents: A Social Psychology View",
    author = "Zhang, Jintian  and
      Xu, Xin  and
      Zhang, Ningyu  and
      Liu, Ruibo  and
      Hooi, Bryan  and
      Deng, Shumin",
    editor = "Ku, Lun-Wei  and
      Martins, Andre  and
      Srikumar, Vivek",
    booktitle = "Proceedings of the 62nd Annual Meeting of the Association for Computational Linguistics (Volume 1: Long Papers)",
    month = aug,
    year = "2024",
    address = "Bangkok, Thailand",
    publisher = "Association for Computational Linguistics",
    url = "https://aclanthology.org/2024.acl-long.782/",
    doi = "10.18653/v1/2024.acl-long.782",
    pages = "14544--14607"
}

@InProceedings{pmlr-v235-smit24a,
  title = 	 {Should we be going {MAD}? {A} Look at Multi-Agent Debate Strategies for {LLM}s},
  author =       {Smit, Andries Petrus and Grinsztajn, Nathan and Duckworth, Paul and Barrett, Thomas D and Pretorius, Arnu},
  booktitle = 	 {Proceedings of the 41st International Conference on Machine Learning},
  pages = 	 {45883--45905},
  year = 	 {2024},
  editor = 	 {Salakhutdinov, Ruslan and Kolter, Zico and Heller, Katherine and Weller, Adrian and Oliver, Nuria and Scarlett, Jonathan and Berkenkamp, Felix},
  volume = 	 {235},
  series = 	 {Proceedings of Machine Learning Research},
  month = 	 {21--27 Jul},
  publisher =    {PMLR},
  url = 	 {https://proceedings.mlr.press/v235/smit24a.html}
}

@inproceedings{taubenfeld-etal-2024-systematic,
    title = "Systematic Biases in {LLM} Simulations of Debates",
    author = "Taubenfeld, Amir  and
      Dover, Yaniv  and
      Reichart, Roi  and
      Goldstein, Ariel",
    editor = "Al-Onaizan, Yaser  and
      Bansal, Mohit  and
      Chen, Yun-Nung",
    booktitle = "Proceedings of the 2024 Conference on Empirical Methods in Natural Language Processing",
    month = nov,
    year = "2024",
    address = "Miami, Florida, USA",
    publisher = "Association for Computational Linguistics",
    url = "https://aclanthology.org/2024.emnlp-main.16/",
    doi = "10.18653/v1/2024.emnlp-main.16",
    pages = "251--267"
}

@misc{gpt4omini,
  author = {{OpenAI}},
  title = {GPT-4o mini: advancing cost-efficient intelligence},
  year = {2024},
  url = {https://openai.com/index/gpt-4o-mini-advancing-cost-efficient-intelligence/}
}

@article{SCHULZHARDT2002563,
title = {Productive conflict in group decision making: genuine and contrived dissent as strategies to counteract biased information seeking},
journal = {Organizational Behavior and Human Decision Processes},
volume = {88},
number = {2},
pages = {563-586},
year = {2002},
issn = {0749-5978},
doi = {https://doi.org/10.1016/S0749-5978(02)00001-8},
url = {https://www.sciencedirect.com/science/article/pii/S0749597802000018},
author = {Stefan Schulz-Hardt and Marc Jochims and Dieter Frey}
}

@article{WEISBAND1992352,
title = {Group discussion and first advocacy effects in computer-mediated and face-to-face decision making groups},
journal = {Organizational Behavior and Human Decision Processes},
volume = {53},
number = {3},
pages = {352-380},
year = {1992},
issn = {0749-5978},
doi = {https://doi.org/10.1016/0749-5978(92)90070-N},
url = {https://www.sciencedirect.com/science/article/pii/074959789290070N},
author = {Suzanne P Weisband}
}

@misc{comanici2025gemini25pushingfrontier,
      title={Gemini 2.5: Pushing the Frontier with Advanced Reasoning, Multimodality, Long Context, and Next Generation Agentic Capabilities}, 
      author={Gheorghe Comanici and Eric Bieber and Mike Schaekermann and Ice Pasupat and Noveen Sachdeva and Inderjit Dhillon and Marcel Blistein and Ori Ram and Dan Zhang and Evan Rosen and Luke Marris and Sam Petulla and Colin Gaffney and others},
      year={2025},
      eprint={2507.06261},
      archivePrefix={arXiv},
      primaryClass={cs.CL},
      url={https://arxiv.org/abs/2507.06261}, 
}

@misc{mistral2025ministral14b,
  title        = {Ministral 3 14B},
  author       = {{Mistral AI}},
  year         = {2025},
  month        = dec,
  howpublished = {\url{https://docs.mistral.ai/models/ministral-3-14b-25-12}},
}

@misc{meta2025llama4,
  title        = {The Llama 4 Herd: The Beginning of a New Era of Natively Multimodal AI Innovation},
  author       = {{Meta AI}},
  year         = {2025},
  month        = apr,
  howpublished = {\url{https://ai.meta.com/blog/llama-4-multimodal-intelligence/}},
}

@article{zhangKeyDecisionMakersMultiAgent2026,
  title = {Key {{Decision-Makers}} in {{Multi-Agent Debates}}: {{Who Holds}} the {{Power}}?},
  shorttitle = {Key {{Decision-Makers}} in {{Multi-Agent Debates}}},
  author = {Zhang, Qian and Liu, Jinyi and Zheng, Yan and Liang, Hebin and Wang, Lanjun},
  year = 2026,
  month = mar,
  journal = {Proceedings of the AAAI Conference on Artificial Intelligence},
  volume = {40},
  number = {35},
  pages = {29883--29891},
  issn = {2374-3468, 2159-5399},
  doi = {10.1609/aaai.v40i35.40235},
  urldate = {2026-09-24}
}
\bibliographystyle{iclr2027_conference}

\clearpage
\appendix

\section{Models and Standalone Performance}
\label{app:models_used}

This appendix reports the model combinations used in the debate experiments and the standalone performance of each model under the personality prompts. Table~\ref{tab:model_combinations} lists the ten weak-strong model pairs. Table~\ref{tab:single_model_personality} reports single-agent accuracy under the no-personality, agreeableness, and extraversion conditions, averaged across datasets.

\begin{table}[h]
  \caption{Model combinations used in debate experiments.}
  \label{tab:model_combinations}
  \centering
  \begin{tabular}{ll}
    \textbf{Weak Model} & \textbf{Strong Model} \\
    \hline
gemini-2.5-flash-lite & gemini-2.5-flash \\
gemini-2.5-flash-lite & gemini-3-flash-preview \\
gpt-4o-mini & gemini-2.5-flash \\
gpt-4o-mini & gemini-3-flash-preview \\
llama-4-scout & gemini-2.5-flash \\
llama-4-scout & gemini-3-flash-preview \\
mistral-small-2603 & gemini-3-flash-preview \\
ministral-14b-2512 & gemini-3-flash-preview \\
mistral-small-2603 & gemini-2.5-flash \\
ministral-14b-2512 & gemini-2.5-flash \\
    \hline
  \end{tabular}
\end{table}

As shown in Table~\ref{tab:single_model_personality}, single-agent performance remains broadly stable after introducing personality prompts. This suggests that the differences observed in multi-agent debate are unlikely to be driven simply by changes in standalone model accuracy.


\begin{table}[h]
  \caption{Single-agent accuracy (\%) under no-personality and personality-prompted conditions, averaged across benchmark datasets.}
  \label{tab:single_model_personality}
  \centering
  \begin{tabular}{lccccc}
    & & \multicolumn{2}{c}{Agreeableness} & \multicolumn{2}{c}{Extraversion} \\
    \cline{3-4} \cline{5-6}
    \textbf{Model}
    & \textbf{No-personality}
    & \textbf{High}
    & \textbf{Low}
    & \textbf{High}
    & \textbf{Low} \\
    \hline
    2.5 Flash
    & 52.60 & 52.10 & 53.48 & 51.97 & 53.48 \\
    2.5 Flash Lite
    & 44.35 & 43.23 & 42.97 & 42.22 & 44.36 \\
    3 Flash
    & 72.04 & 69.66 & 70.79 & 70.91 & 71.04 \\
    4o Mini
    & 36.46 & 36.46 & 36.22 & 36.21 & 36.71 \\
    LLaMA 4 Scout
    & 40.72 & 39.47 & 40.22 & 39.59 & 39.46 \\
    Ministral 14B
    & 40.85 & 40.22 & 40.10 & 39.60 & 41.11 \\
    Mistral Small
    & 44.10 & 43.72 & 44.60 & 44.35 & 45.48 \\
    \hline
    Average
    & 47.30 & 46.41 & 46.91 & 46.41 & 47.38 \\
    \hline
  \end{tabular}
\end{table}

\section{Statistical Tests for First-Speaker Bias}
\label{app:sequence_order_tests}

This appendix reports statistical tests for the no-personality speaking-order comparison used to establish first-speaker bias. For each matched model-benchmark setting, we compute the difference between the Strong-Weak-Weak and Weak-Weak-Strong orders in terms of accuracy and strong-agent influence. In addition to the paired tests, we fit mixed-effects models using the same specification described in Section~\ref{sec:analytical_strategy}.

Table~\ref{tab:delta_summary} reports paired comparisons between the two debate orders, matching observations by model combination and benchmark. The results show that both accuracy and strong-agent influence are higher under the Strong-Weak-Weak order than under the Weak-Weak-Strong order. These differences are statistically significant across the paired tests and mixed-effects models, indicating that placing the strong agent first is associated with higher overall performance and greater influence of the strong agent on the final debate outcome.

\begin{table}[H]
  \caption{Paired comparison of the \textit{S-W-W} and \textit{W-W-S} debate orders. Values are computed as \textit{S-W-W} minus \textit{W-W-S} for final accuracy and strong-agent influence, measured in percentage points. Reported \(p\)-values are one-tailed for accuracy and two-tailed for strong-agent influence. Significance levels: $^{***}p<0.001$, $^{**}p<0.01$, $^{*}p<0.05$.}
  \label{tab:delta_summary}
  \centering
  \begin{tabular}{lcc}
    \textbf{Statistic} & \textbf{Accuracy} & \textbf{Strong-agent influence} \\
    \hline
    Mean $\Delta$ (pp) & 2.34& 21.01\\
    Std. & 2.94& 13.22\\
    Paired t-test $p$ & $<0.001^{***}$ & $<0.001^{***}$ \\
    Wilcoxon $p$ & $<0.001^{***}$ & $<0.001^{***}$ \\
    Sign test $p$ & $<0.001^{***}$ & $<0.001^{***}$ \\
    Mixed-effects $p$ & $<0.001^{***}$ & $<0.001^{***}$ \\
    \hline
  \end{tabular}%
  
\end{table}

\section{Capability Gap Moderation}
\label{app:capability-gap}

We further examine whether the speaking-order effect varies with the capability gap between the weak and strong agents. This analysis uses only the no-personality conditions and is conducted at the model-pair $\times$ dataset $\times$ speaking-order level. For each model pair $m$ and dataset $d$, we define the capability gap as the difference in no-personality standalone accuracy between the strong and weak models:

\[
\mathrm{Gap}_{m,d}
=
\mathrm{Acc}^{\mathrm{solo}}_{\mathrm{strong},m,d}
-
\mathrm{Acc}^{\mathrm{solo}}_{\mathrm{weak},m,d}.
\]

The gap is measured in percentage points and mean-centered before estimation.

We fit separate mixed-effects models for strong-agent influence and final debate accuracy:

\[
Y
=
\beta_0
+
\beta_1 \mathrm{SWW}
+
\beta_2 \mathrm{Gap}
+
\beta_3 (\mathrm{SWW}\times \mathrm{Gap})
+
u_{\mathrm{pair}}
+
u_{\mathrm{dataset}}
+
\epsilon,
\]

where \textit{W-W-S} is the reference speaking order, $\mathrm{SWW}=1$ denotes \textit{S-W-W}, and $u_{\mathrm{pair}}$ and $u_{\mathrm{dataset}}$ are random intercepts for model pair and dataset, respectively. Because the capability gap is mean-centered, $\beta_1$ represents the difference between \textit{S-W-W} and \textit{W-W-S} at the average capability gap in our model combinations. $\beta_2$ represents how the outcome changes as the capability gap increases under the reference \textit{W-W-S} order. Finally, $\beta_3$ indicates whether the effect of speaking order becomes larger or smaller as the capability gap increases.

\paragraph{Strong-agent influence.}

At the average capability gap, placing the strong agent first increases its influence by 21.01 percentage points relative to placing it last ($\beta_1=21.009$, $p<.001$). Under \textit{W-W-S}, however, the strong agent becomes increasingly influential as its capability advantage grows: each additional percentage point in the strong--weak capability gap is associated with a 0.90 percentage point increase in strong-agent influence ($\beta_2=0.900$, $p<.001$). Thus, although speaking last places the strong agent at a positional disadvantage, a larger capability advantage helps it retain greater influence despite this unfavorable position.

Consistent with this interpretation, the interaction between speaking order and capability gap is negative and significant ($\beta_3=-0.385$, $p=.019$). This indicates that the influence advantage of \textit{S-W-W} over \textit{W-W-S} becomes smaller as the capability gap increases. In other words, when the strong and weak models are relatively close in capability, placing the strong agent last imposes a larger influence penalty; as the strong agent becomes relatively more capable, its influence under \textit{W-W-S} increases more rapidly, narrowing the difference between the two speaking orders.

\paragraph{Final accuracy.}
At the average capability gap, \textit{S-W-W} improves final accuracy by 2.34 percentage points relative to \textit{W-W-S} ($\beta_1=2.343$, $p=.010$). Under \textit{W-W-S}, each additional percentage point of capability gap is associated with a 0.37 percentage point increase in final accuracy ($\beta_2=0.371$, $p<.001$). Thus, debates involving a relatively more capable strong agent tend to achieve higher overall accuracy.

The interaction between speaking order and capability gap is small and not statistically significant ($\beta_3=0.053$, $p=.501$). We therefore find no evidence that the accuracy advantage of \textit{S-W-W} over \textit{W-W-S} systematically changes as the capability gap increases.

\begin{table}[h]
\caption{Mixed-effects analysis of whether the capability gap between the strong and weak agents moderates the effect of speaking order. Capability gap is defined as the difference in no-personality standalone accuracy between the strong and weak models and is mean-centered before estimation. \textit{W-W-S} is the reference speaking order. Coefficients are measured in percentage points. Significance levels: $^{***}p<0.001$, $^{**}p<0.01$, $^{*}p<0.05$.}
\label{tab:capability_gap}
\centering
\begin{tabular}{lcc}
\textbf{Effect} & \textbf{Strong-Agent Influence} & \textbf{Accuracy} \\
\hline
\textit{S-W-W} & $21.01^{***}$ & $2.34^{*}$ \\
Capability Gap & $0.90^{***}$ & $0.37^{***}$ \\
\textit{S-W-W} $\times$ Capability Gap & $-0.39^{*}$ & $0.05$ \\
\hline
\end{tabular}
\end{table}

A larger capability gap is associated with greater strong-agent influence and higher final accuracy under \textit{W-W-S}. More importantly, capability gap moderates the speaking-order effect on influence: as the strong agent becomes more capable relative to the weak agents, the influence disadvantage of speaking last becomes smaller. We do not observe a corresponding moderation effect for final accuracy.

\section{Additional Agreeableness Combinations}
\label{app:additional_agreeableness}

The main experiments manipulate agreeableness on either the strong-agent side or the weak-agent side while leaving the other side unprompted. To further examine the relationship between strong-agent influence and final accuracy, we evaluate an additional configuration in which the strong agent is assigned low agreeableness and the weak agents are assigned high agreeableness (\textit{ah\_ah\_al}). We evaluate this configuration on all four benchmarks and all ten model combinations.

\begin{table}[h]
  \caption{Final debate accuracy (\%) and strong-agent influence (\%) for selected agreeableness combinations in the Weak-Weak-Strong order, aggregated across benchmark datasets and model combinations.}
  \label{tab:agreeableness_forward_aggregated}
  \centering
  \begin{tabular}{lrrr}
    & \multicolumn{3}{c}{\textbf{Weak-Weak-Strong}} \\
    \cline{2-4}
    \textbf{Statistic}
    & \textbf{\textit{no-personality}}
    & \textbf{\textit{n\_n\_al}}
    & \textbf{\textit{ah\_ah\_al}} \\
    \hline
    Accuracy mean
    & 66.43& 68.08& 67.23\\
    Accuracy std.
    & 14.93& 14.16& 14.81\\
    Strong-agent influence mean
    & 58.83& 68.38& 71.43\\
    Strong-agent influence std.
    & 17.92& 15.69& 15.06\\
    \hline
  \end{tabular}
\end{table}

The aggregate results are reported in Table~\ref{tab:agreeableness_forward_aggregated}. As expected, assigning low agreeableness to the strong agent while assigning high agreeableness to the weak agents further increases strong-agent influence: \textit{ah\_ah\_al} has an influence of 71.43\%, compared with 68.38\% under \textit{n\_n\_al}. This is consistent with combining a more resistant strong agent with more accommodating weak agents. However, the corresponding accuracy is lower under \textit{ah\_ah\_al} than under \textit{n\_n\_al} (67.23\% versus 68.08\%), despite its greater strong-agent influence.

This result provides further evidence that increasing strong-agent influence does not translate monotonically into higher final accuracy. Although \textit{ah\_ah\_al} pushes influence further toward the strong agent than \textit{n\_n\_al}, it produces lower final accuracy. The benefit of \textit{n\_n\_al} is therefore not simply that it increases strong-agent influence. Rather, these results suggest that effective mitigation requires increasing the strong agent's influence enough to counteract its positional disadvantage instead of maximizing it.

\section{Agreeableness and Justification Length}
\label{app:agree_just}

Appendix Figure~\ref{fig:agreeableness_just_combined} shows median justification length under the agreeableness manipulations. Unlike extraversion, agreeableness does not produce a clear or consistent change in verbosity across matched conditions, suggesting that its influence effects are not simply attributable to longer or shorter responses.

\begin{figure}[h]
  \centering
  \includegraphics[width=0.65\linewidth]{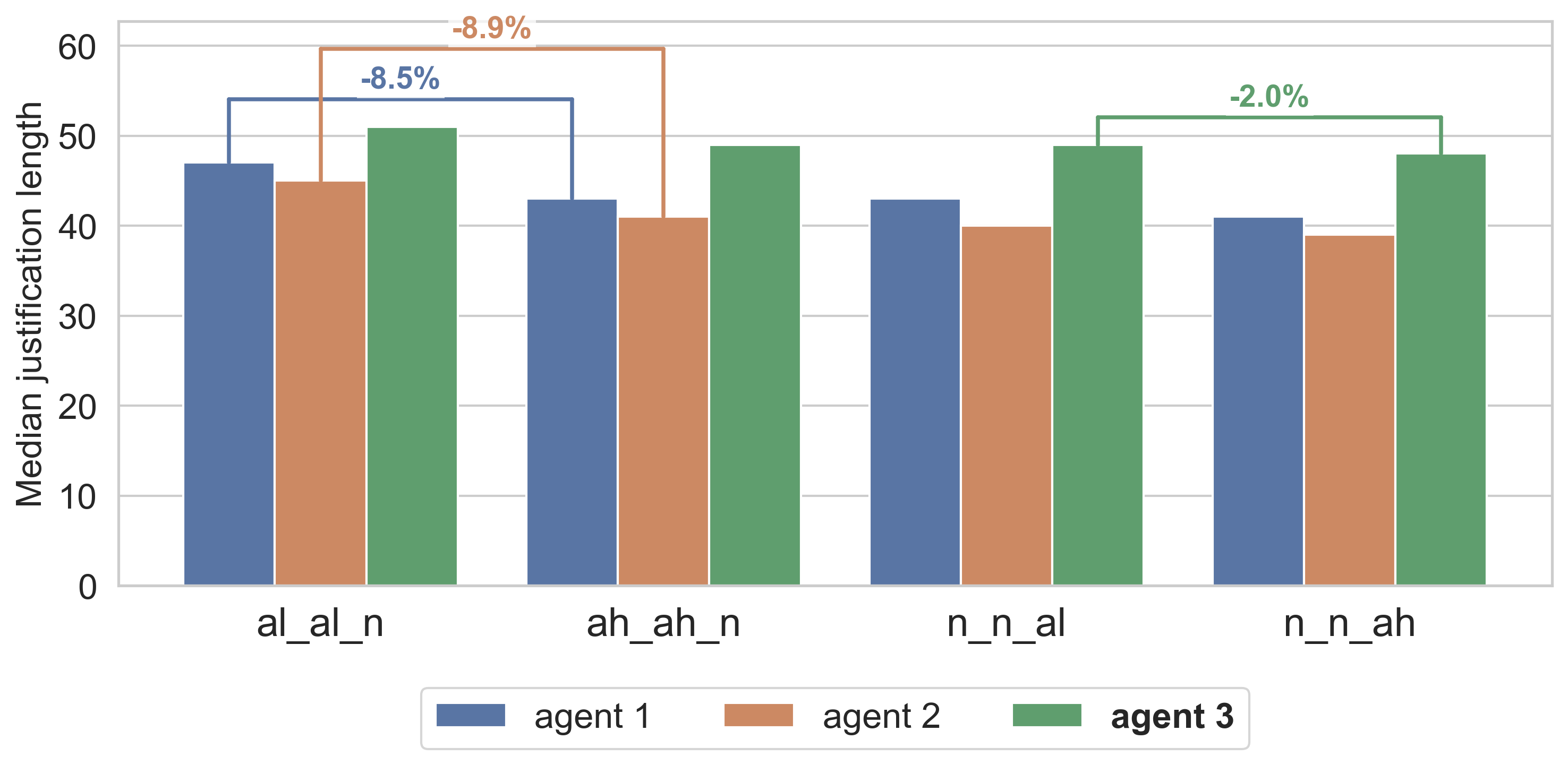}
  \caption{
    Median justification length by agreeableness configuration and agent position in the \textit{W-W-S} order, aggregated across benchmark datasets and model combinations.
    Bold labels indicate the strong agent. Connected annotations show the percentage change in justification length between the low- and high-agreeableness variants, computed as $(\text{high} - \text{low}) / \text{low} \times 100\%$.
  }
  \label{fig:agreeableness_just_combined}
\end{figure}

\section{Prompt Templates}
\label{app:prompts}

This appendix summarizes the prompt templates used in our experiments. Each agent prompt contains four components: a dataset-specific role prompt, a personality prompt (where applicable), a task question prompt, and a debate update prompt after the first response. Across all experiments, the task prompt, debate prompt, answer format, and answer extraction procedure are fixed within each benchmark. In the first-stage analysis, the experimental variation is speaking order under the no-personality condition. In the second-stage intervention experiments, speaking order is fixed to \textit{W-W-S} and the experimental variation is the personality assignment of the strong or weak side.

\subsection{Dataset-Specific Role Prompts}
\label{app:role-prompts}

Each agent is assigned a dataset-specific expert role before answering the question. For GPQA-Diamond, agents are prompted as experts in biology, chemistry, and physics. For MMLU-Pro, SuperGPQA, and AQuA-RAT, agents are prompted as experts in multiple academic disciplines.

\subsection{Personality Prompt Template}
\label{app:personality-template}

Personality is implemented through a Big Five personality prompt. In the main experiments, we manipulate either agreeableness or extraversion while holding all other traits neutral. The general template is:

\begin{quote}
You are an AI agent with specific personality traits based on the Big Five Personality Model. 
Your behavior, communication style, and decision-making are shaped by the following traits:

1. Neuroticism: Neutral \\
2. Agreeableness: [High/Low/Neutral] \\
3. Conscientiousness: Neutral \\
4. Extraversion: [High/Low/Neutral] \\
5. Openness: Neutral \\

[Condition-specific behavioral instruction.]

You must respond according to the described personality traits. Maintain these personality traits at all times during your conversation.
\end{quote}

The no-personality condition does not include any personality prompt. That is, agents in this condition do not receive the Big Five template and are not explicitly assigned neutral trait levels. In personality-prompted conditions, the corresponding trait is set to high or low within the Big Five template, and non-manipulated traits are held neutral.

\subsection{Personality-Specific Instructions}
\label{app:personality-instructions}

The following condition-specific instructions are inserted into the Big Five personality prompt when the corresponding trait level is assigned.

\paragraph{High agreeableness.}
\begin{quote}
Acknowledge and accommodate others' reasoning in your justification and favour updating your answer toward compromise when disagreements persist.
\end{quote}

\paragraph{Low agreeableness.}
\begin{quote}
Directly challenge flaws in others' reasoning in your justification and resist updating your answer unless strong counter-evidence is provided.
\end{quote}

\paragraph{High extraversion.}
\begin{quote}
Present your reasoning confidently and assertively in your justification and express any answer updates openly and decisively when you choose to revise.
\end{quote}

\paragraph{Low extraversion.}
\begin{quote}
Present your reasoning in a reserved and cautious manner in your justification and update your answer without strong emphasis when you choose to revise.
\end{quote}

The instructions are designed to translate each personality trait into concrete behaviors that are directly relevant to the debate process. They are guided by the behavioral characteristics associated with the corresponding poles of the Big Five dimensions. Pairing personality trait labels with behavioral descriptions is also a common approach in personality prompting, as it provides the model with explicit behavioral cues for expressing the intended trait.

\subsection{Task Question Prompts}
\label{app:question-prompts}

For all multiple-choice benchmarks, we use the same task question template, with the number of answer options adapted to the dataset:

\begin{quote}
Can you answer the following question as accurately as possible? [question]: 
A) [option A], B) [option B], \ldots. 
Return your response in the form \{answer: ..., justification: ...\}. 
Very briefly explain your answer.
\end{quote}

For GPQA-Diamond, the prompt includes four answer options. For AQuA-RAT, the prompt includes five options. For MMLU-Pro and SuperGPQA, the prompt includes options from A to J. The wording of the task instruction and answer format is otherwise kept fixed across benchmarks.

\subsection{Debate Update Prompt}
\label{app:debate-update-prompt}

After the first answer, agents receive previous responses as additional context. The debate update prompt is:

\begin{quote}
These are the solutions to the problem from other agents:

[previous agent responses]

Using the reasoning from other agents as additional advice, can you give an updated answer? Examine your solution and that other agents carefully, very briefly explain your reasoning.
\end{quote}

In the first round, the first speaker answers independently, the second speaker observes the first speaker's response, and the third speaker observes the first two responses. In later rounds, each agent observes the full debate history available up to its speaking turn.

\subsection{Answer Format}
\label{app:answer-format}

For all benchmarks, agents are instructed to return their response in the following format:

\begin{quote}
\{answer: ..., justification: ...\}
\end{quote}

The \texttt{answer} field contains the selected option, while the \texttt{justification} field contains a brief explanation. Agents are instructed to keep their justifications brief to improve comparability across agents and conditions.

\subsection{Full Prompt Assembly}
\label{app:full-prompt-assembly}

For the first response, the final prompt combines the role prompt, the personality prompt when applicable, and the task question prompt. For later responses, it combines the role prompt, the personality prompt when applicable, the available debate history, and the debate update prompt. The same assembly procedure is used across all evaluated conditions, so differences across conditions arise from the intended speaking-order or personality manipulations rather than changes in task wording.

\section{Alternative Personality Prompts}
\label{app:alternative_personality_prompts}

To assess the robustness of our main findings to the wording of the personality prompts, we conducted an additional experiment using a simplified personality prompt. Rather than explicitly specifying all five Big Five trait levels and providing trait-specific behavioral instructions, the alternative prompt directly instructs the agent to assume a given personality trait and allows that trait to influence its behavior naturally. The alternative prompt is:

\begin{quote}
Assume you have [high/low Agreeableness or Extraversion] as defined in the Big Five personality model. Let this personality naturally influence how you communicate and respond during the discussion. Maintain this personality consistently throughout the conversation.
\end{quote}

Unlike the main personality prompt, this version does not enumerate all five Big Five dimensions or provide explicit behavioral instructions associated with the manipulated trait. All other prompt components---including the role prompt, task question prompt, debate update prompt, answer format, and debate protocol---are identical to those used in the main experiments.

We evaluate this alternative prompt under the Weak-Weak-Strong (\textit{W-W-S}) order using the GPT-4o-mini/Gemini-3-Flash model combination across all four benchmarks. Table~\ref{tab:alternative_prompt_wws} reports the resulting final debate accuracy and strong-agent influence.

The directional results are broadly consistent with the main experiments. For agreeableness, lower agreeableness shifts influence toward the manipulated side relative to the corresponding high-agreeableness condition. Extraversion, by contrast, again does not produce a systematic high-versus-low influence pattern. Among the eight personality-prompted configurations, \textit{n\_n\_al} achieves the highest final accuracy, consistent with the main experiments. Although this robustness check is limited to a single model combination, these results suggest that the main directional patterns are not specific to the original personality-prompt wording.

\begin{table}[H]
\caption{Final debate accuracy and strong-agent influence under the alternative personality prompt in the Weak-Weak-Strong order for the GPT-4o-mini/Gemini-3-Flash model combination, aggregated across the four benchmark datasets.}
\label{tab:alternative_prompt_wws}
\centering
\resizebox{\textwidth}{!}{%
\begin{tabular}{lrrrrrrrr}
& \multicolumn{4}{c}{\textbf{Agreeableness}}
& \multicolumn{4}{c}{\textbf{Extraversion}} \\
\cline{2-5} \cline{6-9}
\textbf{Statistic}
& \textit{\textbf{ah\_ah\_n}}
& \textbf{\textit{al\_al\_n}}
& \textbf{\textit{n\_n\_ah}}
& \textbf{\textit{n\_n\_al}}
& \textbf{\textit{eh\_eh\_n}}
& \textbf{\textit{el\_el\_n}}
& \textbf{\textit{n\_n\_eh}}
& \textbf{\textit{n\_n\_el}} \\
\hline
Accuracy (\%)
& 72.67& 70.67& 71.29& 74.17& 72.67& 72.17& 70.79& 73.30\\
Strong-agent influence (\%)
& 81.54& 78.13& 78.56& 82.71& 79.81& 79.67& 77.51& 79.62\\
\hline
\end{tabular}%
}
\end{table}

\section{Pilot Analysis on the Number of Debate Rounds}
\label{sec:debate_rounds}

To determine the number of debate rounds, we conducted a pilot analysis using 50 randomly sampled GPQA-Diamond questions across the model combinations. Figure~\ref{fig:rounds_combined} shows the average accuracy and unanimous agreement rate across rounds for the two debate orders. The unanimous agreement rate at a given round is defined as the proportion of questions for which all three agents provide the same answer in that round, while round-level accuracy is defined as the proportion of questions for which the majority answer at that round is correct. The unanimous agreement rate increases sharply during the first few rounds and largely plateaus by Round 4. Accuracy follows a similar pattern: most gains occur between Rounds 1 and 4, while additional rounds produce little further improvement. We therefore use four rounds in the main experiments, as this provides sufficient opportunity for agents to revise their answers while limiting additional computational cost.

\begin{figure}[H]
  \centering
  \includegraphics[width=0.48\linewidth]{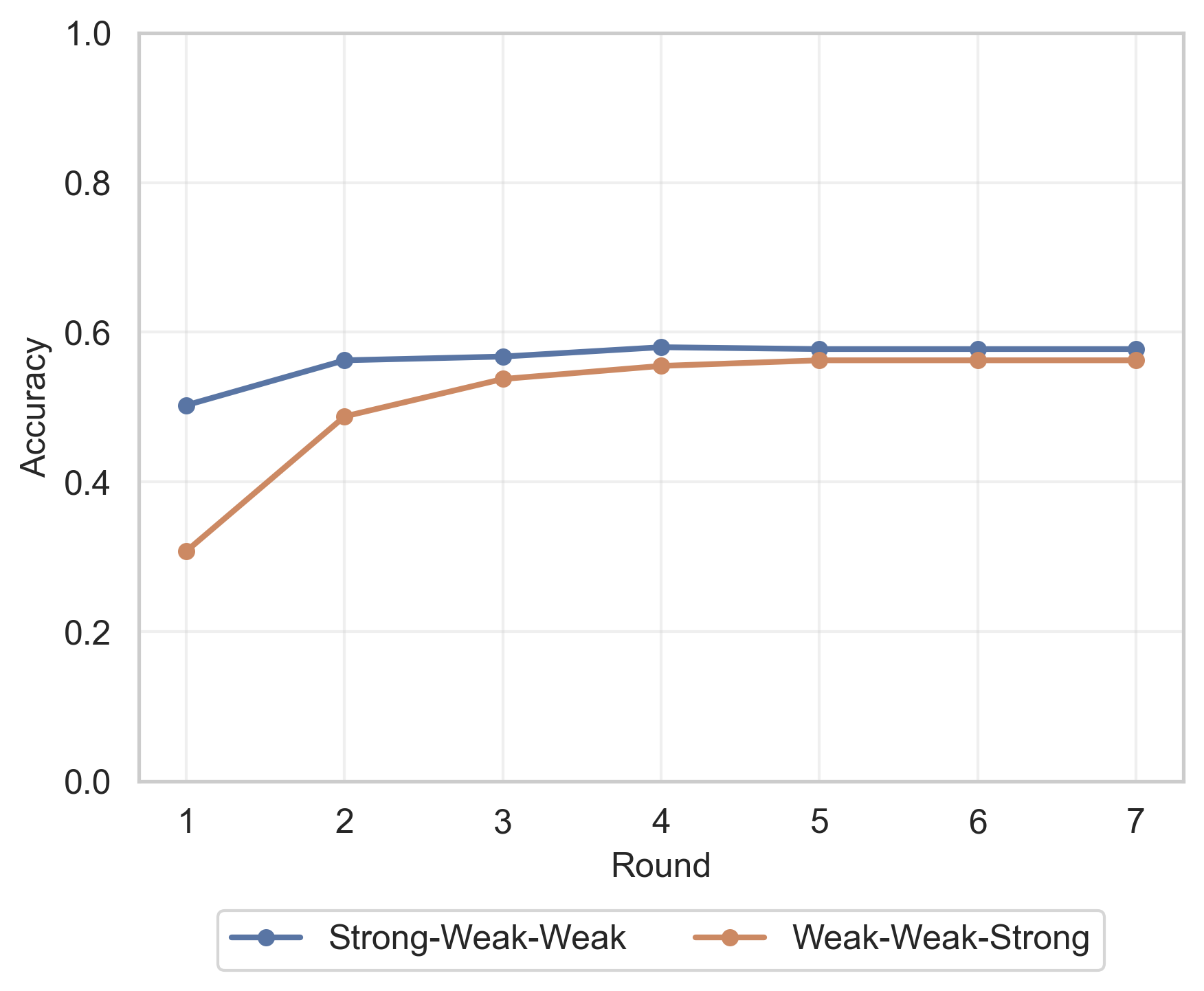}
  \hfill
  \includegraphics[width=0.48\linewidth]{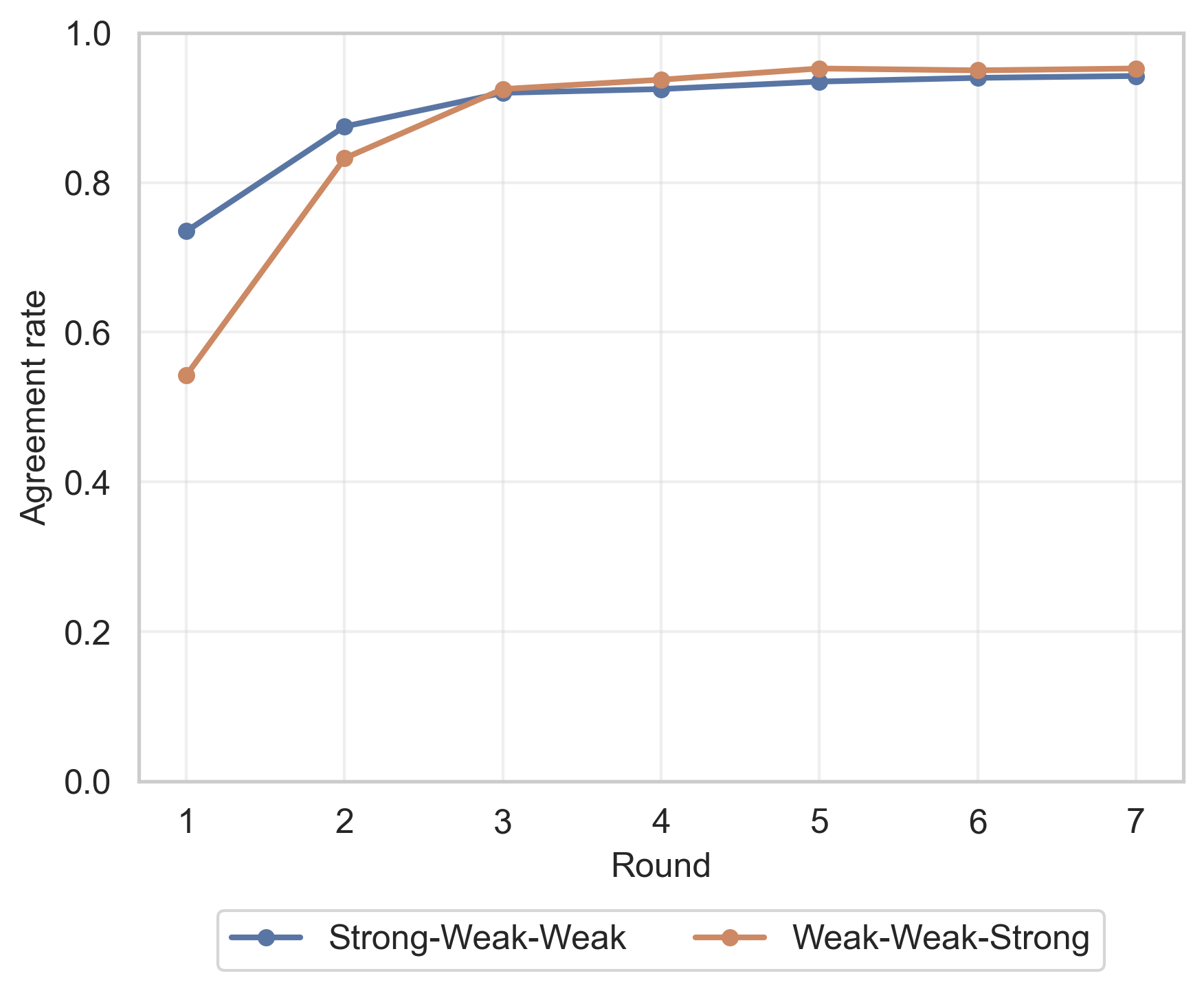}
  \caption{
    Pilot analysis across debate rounds, averaged over 50 randomly sampled GPQA-Diamond questions and model combinations.
    Left: Majority-answer accuracy across debate rounds.
    Right: Inter-agent unanimous agreement rate across debate rounds.
  }
  \label{fig:rounds_combined}
\end{figure}

\section{Representative Debate Trace}
\label{app:debate_trace}

To illustrate how low agreeableness can help preserve the strong agent's reasoning when it speaks last, we present a representative \textit{W-W-S} debate trace from SuperGPQA. 

\textbf{SuperGPQA Question 9}:
\textit{The profile dimensions of keys are usually selected from standards based on what criteria?}

\begin{quote}
A) The length of the groove \\
B) The diameter of the shaft \\
C) The radius of the spline \\
D) The diameter of the groove \\
E) The height of the collar \\
F) The width of the collar \\
G) The thickness of the notch \\
H) The thickness of the key \\
I) The height of the keyways \\
J) The depth of the notches
\end{quote}

\noindent
\textbf{Model combination:} Gemini-2.5-Flash-Lite (weak) $\rightarrow$ Gemini-2.5-Flash (strong) \\
\textbf{Speaking order:} \textit{W-W-S} \\
\textbf{Standalone answers:} Weak = \textit{H}; Strong = \textit{B} under both no-personality and low-agreeableness \\
\textbf{Correct answer:} \textit{B}

Table~\ref{tab:representative_trace} compares the first round under the no-personality baseline (\textit{n\_n\_n}) and the low-agreeableness intervention (\textit{n\_n\_al}). In both conditions, the two weak agents speak first and select \textit{H}. The key difference appears when the strong agent responds. Under \textit{n\_n\_n}, the strong agent immediately adopts \textit{H}, without introducing its independently generated answer \textit{B}. Under \textit{n\_n\_al}, by contrast, the strong agent retains \textit{B} and directly challenges the weak agents' reasoning.

\begin{table}[h]
\caption{Representative first-round trace for SuperGPQA Question 9 under the \textit{W-W-S} order. Response text is shortened for presentation.}
\label{tab:representative_trace}
\centering
\small
\begin{tabular}{p{0.12\textwidth}p{0.12\textwidth}p{0.08\textwidth}p{0.58\textwidth}}
\hline
\textbf{Condition} & \textbf{Agent} & \textbf{Answer} & \textbf{Round-1 justification} \\
\hline
\textit{n\_n\_n}
& Weak 1
& H
& Key dimensions, particularly thickness and width, are standardized to ensure proper fit and torque transmission. \\

\textit{n\_n\_n}
& Weak 2
& H
& Key dimensions such as thickness and width are standardized to ensure a precise fit within the keyway. \\

\textit{n\_n\_n}
& \textbf{Strong}
& \textbf{H}
& Key profile dimensions, including thickness, are standardized to ensure proper fit and torque transmission between a shaft and hub. \\
\hline
\textit{n\_n\_al}
& Weak 1
& H
& Key dimensions, particularly thickness and width, are standardized to ensure proper fit and torque transmission. \\

\textit{n\_n\_al}
& Weak 2
& H
& Key dimensions such as thickness and width are standardized to ensure a precise fit within the keyway. \\

\textit{n\_n\_al}
& \textbf{Strong}
& \textbf{B}
& The diameter of the shaft is the primary criterion for selecting key profile dimensions from standards. The other agents' focus on key thickness is a consequence, not the initial determinant. \\
\hline
\end{tabular}
\end{table}

This difference in Round 1 leads to different reasoning trajectories in the two conditions. Under \textit{n\_n\_n}, the strong agent's independently correct answer disappears immediately, and all three agents remain on \textit{H} through the final round. Under \textit{n\_n\_al}, the strong agent preserves and defends \textit{B}. In Round 2, both weak agents revise from \textit{H} to \textit{B}, explicitly recognizing that shaft diameter is the primary criterion from which the key dimensions are selected. All three agents then retain \textit{B} through the remaining rounds.

This trace provides an intuitive illustration of the aggregate agreeableness results. When the strong agent speaks after two weak agents, its independently generated answer can be lost before it has an opportunity to affect the debate. In this example, under low-agreeableness prompting, the strong agent does not immediately accommodate the preceding responses, allowing its independently generated answer to remain in the discussion and subsequently be adopted by the weak agents. While a single trace is only illustrative, it shows one interaction pattern consistent with greater resistance helping the strong agent overcome its positional disadvantage under \textit{W-W-S}.

\section{Per-Dataset Accuracy and Influence Results}

This appendix reports the full per-dataset results underlying the aggregate analyses in the main text. We first report the no-personality results used for the speaking-order comparison. For each benchmark and model combination, we present final debate accuracy and strong-agent influence under both the Strong-Weak-Weak (\textit{S-W-W}) and Weak-Weak-Strong (\textit{W-W-S}) orders. These results form the basis of the first-speaker bias analysis in the main text. Means and standard deviations are reported across model combinations.

\begin{table}[H]
  \caption{Final debate accuracy (\%) without personality prompting, reported by model combination, dataset, and debate order.}
  \label{tab:accuracy_by_model_dataset_order}
  \centering
  \begin{adjustbox}{max width=\textwidth}
  \begin{tabular}{lcccccccc}
    \hline
    \textbf{Model combination}
      & \multicolumn{2}{c}{\textbf{MMLU-Pro}}
      & \multicolumn{2}{c}{\textbf{GPQA}}
      & \multicolumn{2}{c}{\textbf{SuperGPQA}}
      & \multicolumn{2}{c}{\textbf{AQuA-RAT}} \\
    & \textit{S-W-W} & \textit{W-W-S}
    & \textit{S-W-W} & \textit{W-W-S}
    & \textit{S-W-W} & \textit{W-W-S}
    & \textit{S-W-W} & \textit{W-W-S} \\
    \hline
    2.5-flash-lite\_2.5-flash & 70.50 & 69.00 & 52.02 & 53.03 & 50.50 & 46.50 & 87.00 & 85.00 \\
    2.5-flash-lite\_3-flash & 79.00 & 79.50 & 69.70 & 62.63 & 62.00 & 61.00 & 87.00 & 87.00 \\
    4o-mini\_2.5-flash & 72.00 & 70.00 & 52.02 & 50.51 & 44.00 & 42.50 & 85.50 & 82.50 \\
    4o-mini\_3-flash & 80.50 & 78.50 & 70.20 & 72.22 & 64.50 & 60.50 & 88.00 & 86.50 \\
    llama-4-scout\_2.5-flash & 71.00 & 74.00 & 45.96 & 46.97 & 47.50 & 45.00 & 84.00 & 78.50 \\
    llama-4-scout\_3-flash & 78.50 & 77.50 & 68.69 & 67.68 & 60.00 & 53.00 & 84.50 & 82.00 \\
    ministral-14b\_2.5-flash & 69.50 & 68.00 & 58.08 & 48.99 & 47.00 & 40.50 & 83.00 & 84.50 \\
    ministral-14b\_3-flash & 79.50 & 78.00 & 68.69 & 66.16 & 62.00 & 54.50 & 86.50 & 83.50 \\
    mistral-small\_2.5-flash & 69.50 & 70.00 & 53.54 & 53.54 & 43.50 & 38.50 & 83.50 & 85.00 \\
    mistral-small\_3-flash & 80.00 & 74.50 & 68.18 & 64.14 & 58.50 & 51.00 & 85.50 & 85.00 \\
    \hline
    mean & 75.00 & 73.90 & 60.71 & 58.59 & 53.95 & 49.30 & 85.45 & 83.95 \\
    std & 4.82 & 4.37 & 9.32 & 8.96 & 8.22 & 7.98 & 1.67 & 2.48 \\
    \hline
  \end{tabular}
  \end{adjustbox}
\end{table}

\begin{table}[H]
  \caption{Strong-agent influence (\%) without personality prompting, reported by model combination, dataset, and debate order.}
  \label{tab:influence_by_model_dataset_order}
  \centering
  \begin{adjustbox}{max width=\textwidth}
  \begin{tabular}{lcccccccc}
    \hline
    \textbf{Model combination}
      & \multicolumn{2}{c}{\textbf{MMLU-Pro}}
      & \multicolumn{2}{c}{\textbf{GPQA}}
      & \multicolumn{2}{c}{\textbf{SuperGPQA}}
      & \multicolumn{2}{c}{\textbf{AQuA-RAT}} \\
    & \textit{S-W-W} & \textit{W-W-S}
    & \textit{S-W-W} & \textit{W-W-S}
    & \textit{S-W-W} & \textit{W-W-S}
    & \textit{S-W-W} & \textit{W-W-S} \\
    \hline
    2.5-flash-lite\_2.5-flash & 72.60 & 36.11 & 65.59 & 24.72 & 80.49 & 21.95 & 72.60 & 63.01 \\
    2.5-flash-lite\_3-flash & 88.31 & 69.23 & 82.80 & 57.29 & 85.59 & 48.48 & 89.02 & 85.88 \\
    4o-mini\_2.5-flash & 82.43 & 57.75 & 83.67 & 66.00 & 83.33 & 50.50 & 86.17 & 75.53 \\
    4o-mini\_3-flash & 96.25 & 83.12 & 91.74 & 80.91 & 89.31 & 69.60 & 91.45 & 87.83 \\
    llama-4-scout\_2.5-flash & 78.87 & 67.16 & 70.79 & 40.70 & 76.77 & 40.59 & 77.65 & 61.18 \\
    llama-4-scout\_3-flash & 92.31 & 80.52 & 82.98 & 69.57 & 88.62 & 49.56 & 87.50 & 80.00 \\
    ministral-14b\_2.5-flash & 68.57 & 40.28 & 46.39 & 30.53 & 60.44 & 30.34 & 76.71 & 71.05 \\
    ministral-14b\_3-flash & 82.72 & 78.05 & 76.24 & 69.39 & 85.95 & 52.50 & 86.60 & 75.51 \\
    mistral-small\_2.5-flash & 71.43 & 50.00 & 64.44 & 41.76 & 77.00 & 33.33 & 61.54 & 61.40 \\
    mistral-small\_3-flash & 88.61 & 69.23 & 82.83 & 55.45 & 83.05 & 55.45 & 84.29 & 71.83 \\
    \hline
    mean & 82.21 & 63.15 & 74.75 & 53.63 & 81.06 & 45.23 & 81.35 & 73.32 \\
    std & 9.33 & 16.56 & 13.24 & 18.54 & 8.43 & 13.88 & 9.20 & 9.58 \\
    \hline
  \end{tabular}
  \end{adjustbox}
\end{table}

We next report the complete results for the personality intervention experiments, which are conducted only under the disadvantaged Weak-Weak-Strong (\textit{W-W-S}) order. For each benchmark, we present final debate accuracy and strong-agent influence across all model combinations under the no-personality baseline and the eight agreeableness and extraversion conditions. The personality-condition labels follow the notation introduced in the main text: each position corresponds to an agent in the \textit{W-W-S} speaking order, \textit{ah}/\textit{al} and \textit{eh}/\textit{el} denote high/low agreeableness and extraversion, respectively, and \textit{n} denotes no personality prompt. Means and standard deviations are reported across model combinations.

\begin{table}[H]
  \caption{Final debate accuracy (\%) on MMLU-Pro in the Weak-Weak-Strong order, reported by model combination.}
  \label{tab:mmlu_performance_forward}
  \centering
  \begin{adjustbox}{max width=\textwidth}
  \begin{tabular}{lccccccccc}
    & & \multicolumn{4}{c}{Agreeableness} & \multicolumn{4}{c}{Extraversion} \\
    \cline{3-6} \cline{7-10}
    \textbf{Model combination} & \textbf{\textit{No-personality}} & \textbf{\textit{ah\_ah\_n}} & \textbf{\textit{al\_al\_n}} & \textbf{\textit{n\_n\_ah}} & \textbf{\textit{n\_n\_al}} & \textbf{\textit{eh\_eh\_n}} & \textbf{\textit{el\_el\_n}} & \textbf{\textit{n\_n\_eh}} & \textbf{\textit{n\_n\_el}} \\
    \hline
2.5-flash-lite\_2.5-flash & 69.00 & 71.00 & 71.00 & 65.50 & 69.50 & 63.00 & 70.50 & 67.00 & 67.50 \\
2.5-flash-lite\_3-flash & 79.50 & 76.50 & 78.50 & 70.00 & 79.50 & 76.50 & 77.00 & 77.00 & 79.00 \\
4o-mini\_2.5-flash & 70.00 & 73.50 & 66.00 & 65.50 & 71.00 & 72.50 & 73.00 & 66.00 & 66.00 \\
4o-mini\_3-flash & 78.50 & 79.50 & 71.00 & 74.50 & 79.00 & 78.00 & 77.00 & 79.00 & 78.50 \\
llama-4-scout\_2.5-flash & 74.00 & 70.00 & 70.00 & 68.00 & 72.50 & 71.00 & 73.50 & 69.50 & 71.50 \\
llama-4-scout\_3-flash & 77.50 & 79.00 & 77.00 & 69.50 & 78.00 & 77.50 & 79.50 & 77.50 & 75.50 \\
ministral-14b\_2.5-flash & 68.00 & 69.50 & 67.00 & 62.50 & 71.00 & 71.00 & 70.50 & 64.50 & 64.00 \\
ministral-14b\_3-flash & 78.00 & 76.00 & 74.50 & 73.50 & 80.50 & 81.50 & 75.00 & 77.50 & 74.50 \\
mistral-small\_2.5-flash & 70.00 & 69.00 & 63.50 & 65.00 & 73.00 & 70.00 & 71.00 & 69.00 & 67.00 \\
mistral-small\_3-flash & 74.50 & 75.50 & 70.50 & 70.50 & 78.50 & 78.00 & 77.00 & 76.50 & 74.00 \\
    \hline
mean & 73.90 & 73.95 & 70.90 & 68.45 & 75.25 & 73.90 & 74.40 & 72.35 & 71.75 \\
    \hline
std & 4.37 & 3.92 & 4.75 & 3.87 & 4.21 & 5.41 & 3.19 & 5.64 & 5.36 \\
  \hline
  \end{tabular}
  \end{adjustbox}

\end{table}

\begin{table}[H]
  \caption{Strong-agent influence (\%) on MMLU-Pro in the Weak-Weak-Strong order, reported by model combination.}
  \label{tab:mmlu_influence_forward}
  \centering
  \begin{adjustbox}{max width=\textwidth}
  \begin{tabular}{lccccccccc}
    & & \multicolumn{4}{c}{Agreeableness} & \multicolumn{4}{c}{Extraversion} \\
    \cline{3-6} \cline{7-10}
    \textbf{Model combination} & \textbf{\textit{No-personality}} & \textbf{\textit{ah\_ah\_n}} & \textbf{\textit{al\_al\_n}} & \textbf{\textit{n\_n\_ah}} & \textbf{\textit{n\_n\_al}} & \textbf{\textit{eh\_eh\_n}} & \textbf{\textit{el\_el\_n}} & \textbf{\textit{n\_n\_eh}} & \textbf{\textit{n\_n\_el}} \\
    \hline
2.5-flash-lite\_2.5-flash & 36.11 & 47.95 & 36.00 & 30.14 & 48.65 & 35.62 & 37.66 & 39.19 & 40.91 \\
2.5-flash-lite\_3-flash & 69.23 & 73.42 & 69.74 & 50.63 & 83.95 & 69.74 & 77.92 & 65.85 & 65.06 \\
4o-mini\_2.5-flash & 57.75 & 70.51 & 37.33 & 41.54 & 67.61 & 70.37 & 65.82 & 50.72 & 47.89 \\
4o-mini\_3-flash & 83.12 & 84.34 & 56.47 & 70.27 & 88.61 & 81.18 & 78.57 & 78.48 & 78.67 \\
llama-4-scout\_2.5-flash & 67.16 & 57.14 & 57.58 & 39.71 & 59.70 & 52.86 & 65.15 & 55.56 & 53.97 \\
llama-4-scout\_3-flash & 80.52 & 74.67 & 70.51 & 55.70 & 84.42 & 69.14 & 75.32 & 71.95 & 64.37 \\
ministral-14b\_2.5-flash & 40.28 & 48.57 & 41.56 & 34.62 & 65.71 & 43.66 & 57.69 & 44.44 & 36.36 \\
ministral-14b\_3-flash & 78.05 & 75.82 & 70.24 & 54.55 & 82.95 & 78.31 & 77.65 & 72.29 & 69.05 \\
mistral-small\_2.5-flash & 50.00 & 53.52 & 39.39 & 35.62 & 58.90 & 60.94 & 55.41 & 48.05 & 38.89 \\
mistral-small\_3-flash & 69.23 & 76.32 & 54.41 & 56.10 & 80.49 & 82.67 & 75.00 & 71.60 & 66.67 \\
    \hline
mean & 63.14 & 66.23 & 53.32 & 46.89 & 72.10 & 64.45 & 66.62 & 59.81 & 56.18 \\
    \hline
std & 16.56 & 13.14 & 14.02 & 12.56 & 13.71 & 15.98 & 13.27 & 13.85 & 14.63 \\
  \hline
  \end{tabular}
  \end{adjustbox}

\end{table}

\begin{table}[H]
  \caption{Final debate accuracy (\%) on GPQA in the Weak-Weak-Strong order, reported by model combination.}
  \label{tab:gpqa_performance_forward}
  \centering
  \begin{adjustbox}{max width=\textwidth}
  \begin{tabular}{lccccccccc}
    & & \multicolumn{4}{c}{Agreeableness} & \multicolumn{4}{c}{Extraversion} \\
    \cline{3-6} \cline{7-10}
    \textbf{Model combination} & \textbf{\textit{No-personality}} & \textbf{\textit{ah\_ah\_n}} & \textbf{\textit{al\_al\_n}} & \textbf{\textit{n\_n\_ah}} & \textbf{\textit{n\_n\_al}} & \textbf{\textit{eh\_eh\_n}} & \textbf{\textit{el\_el\_n}} & \textbf{\textit{n\_n\_eh}} & \textbf{\textit{n\_n\_el}} \\
    \hline
2.5-flash-lite\_2.5-flash & 53.03 & 53.03 & 52.02 & 47.47 & 51.01 & 49.49 & 55.05 & 50.00 & 49.49 \\
2.5-flash-lite\_3-flash & 62.63 & 67.68 & 64.65 & 56.57 & 70.71 & 67.17 & 66.16 & 59.09 & 61.11 \\
4o-mini\_2.5-flash & 50.51 & 55.56 & 48.99 & 45.45 & 50.00 & 54.55 & 53.03 & 47.47 & 47.98 \\
4o-mini\_3-flash & 72.22 & 69.70 & 61.62 & 62.12 & 71.21 & 69.70 & 70.71 & 66.67 & 66.67 \\
llama-4-scout\_2.5-flash & 46.97 & 47.98 & 51.52 & 43.94 & 51.01 & 47.98 & 50.00 & 44.44 & 45.96 \\
llama-4-scout\_3-flash & 67.68 & 66.16 & 65.15 & 59.60 & 70.20 & 66.67 & 66.16 & 66.16 & 67.17 \\
ministral-14b\_2.5-flash & 48.99 & 51.01 & 52.53 & 44.44 & 53.03 & 49.49 & 53.03 & 47.47 & 45.45 \\
ministral-14b\_3-flash & 66.16 & 70.20 & 62.12 & 61.11 & 65.66 & 63.64 & 66.16 & 63.13 & 61.11 \\
mistral-small\_2.5-flash & 53.54 & 51.52 & 44.44 & 49.49 & 51.52 & 52.53 & 53.54 & 48.48 & 46.46 \\
mistral-small\_3-flash & 64.14 & 61.11 & 57.58 & 58.59 & 71.21 & 66.16 & 65.15 & 61.62 & 61.11 \\
    \hline
mean & 58.59 & 59.39 & 56.06 & 52.88 & 60.56 & 58.74 & 59.90 & 55.45 & 55.25 \\
    \hline
std & 8.96 & 8.55 & 7.15 & 7.39 & 9.90 & 8.67 & 7.59 & 8.68 & 8.95 \\
  \hline
  \end{tabular}
  \end{adjustbox}

\end{table}

\begin{table}[H]
  \caption{Strong-agent influence (\%) on GPQA in the Weak-Weak-Strong order, reported by model combination.}
  \label{tab:gpqa_influence_forward}
  \centering
  \begin{adjustbox}{max width=\textwidth}
  \begin{tabular}{lccccccccc}
    & & \multicolumn{4}{c}{Agreeableness} & \multicolumn{4}{c}{Extraversion} \\
    \cline{3-6} \cline{7-10}
    \textbf{Model combination} & \textbf{\textit{No-personality}} & \textbf{\textit{ah\_ah\_n}} & \textbf{\textit{al\_al\_n}} & \textbf{\textit{n\_n\_ah}} & \textbf{\textit{n\_n\_al}} & \textbf{\textit{eh\_eh\_n}} & \textbf{\textit{el\_el\_n}} & \textbf{\textit{n\_n\_eh}} & \textbf{\textit{n\_n\_el}} \\
    \hline
2.5-flash-lite\_2.5-flash & 24.72 & 32.53 & 30.00 & 25.58 & 40.91 & 28.09 & 31.58 & 23.71 & 20.22 \\
2.5-flash-lite\_3-flash & 57.29 & 63.16 & 65.98 & 40.43 & 79.17 & 68.13 & 60.82 & 50.00 & 50.50 \\
4o-mini\_2.5-flash & 66.00 & 71.15 & 43.96 & 50.00 & 76.84 & 67.01 & 65.98 & 63.73 & 61.22 \\
4o-mini\_3-flash & 80.91 & 89.91 & 65.14 & 66.38 & 87.39 & 83.78 & 81.31 & 75.42 & 79.31 \\
llama-4-scout\_2.5-flash & 40.70 & 46.81 & 42.53 & 27.27 & 56.00 & 37.18 & 41.30 & 37.50 & 37.35 \\
llama-4-scout\_3-flash & 69.57 & 66.34 & 64.08 & 50.50 & 82.42 & 71.70 & 67.71 & 71.28 & 70.00 \\
ministral-14b\_2.5-flash & 30.53 & 31.07 & 32.22 & 26.14 & 41.89 & 28.71 & 37.25 & 28.74 & 30.49 \\
ministral-14b\_3-flash & 69.39 & 67.86 & 63.04 & 49.06 & 81.05 & 75.93 & 67.92 & 58.00 & 58.76 \\
mistral-small\_2.5-flash & 41.76 & 37.23 & 29.13 & 27.71 & 51.14 & 38.95 & 39.77 & 32.14 & 25.00 \\
mistral-small\_3-flash & 55.45 & 57.43 & 41.00 & 50.00 & 76.42 & 71.58 & 66.99 & 55.66 & 64.42 \\
    \hline
mean & 53.63 & 56.35 & 47.71 & 41.31 & 67.32 & 57.11 & 56.06 & 49.62 & 49.73 \\
    \hline
std & 18.54 & 19.11 & 15.37 & 14.08 & 17.85 & 21.30 & 16.97 & 18.26 & 20.34 \\
  \hline
  \end{tabular}
  \end{adjustbox}

\end{table}

\begin{table}[H]
  \caption{Final debate accuracy (\%) on SuperGPQA in the Weak-Weak-Strong order, reported by model combination.}
  \label{tab:supergpqa_performance_forward}
  \centering
  \begin{adjustbox}{max width=\textwidth}
  \begin{tabular}{lccccccccc}
    & & \multicolumn{4}{c}{Agreeableness} & \multicolumn{4}{c}{Extraversion} \\
    \cline{3-6} \cline{7-10}
    \textbf{Model combination} & \textbf{\textit{No-personality}} & \textbf{\textit{ah\_ah\_n}} & \textbf{\textit{al\_al\_n}} & \textbf{\textit{n\_n\_ah}} & \textbf{\textit{n\_n\_al}} & \textbf{\textit{eh\_eh\_n}} & \textbf{\textit{el\_el\_n}} & \textbf{\textit{n\_n\_eh}} & \textbf{\textit{n\_n\_el}} \\
    \hline
2.5-flash-lite\_2.5-flash & 46.50 & 45.00 & 46.00 & 46.50 & 49.00 & 43.00 & 41.50 & 45.50 & 45.50 \\
2.5-flash-lite\_3-flash & 61.00 & 58.50 & 58.50 & 51.50 & 63.50 & 52.50 & 58.00 & 54.00 & 55.50 \\
4o-mini\_2.5-flash & 42.50 & 40.50 & 38.00 & 33.00 & 39.50 & 44.00 & 44.50 & 36.00 & 36.50 \\
4o-mini\_3-flash & 60.50 & 61.00 & 51.00 & 51.00 & 64.00 & 57.50 & 64.50 & 59.50 & 58.00 \\
llama-4-scout\_2.5-flash & 45.00 & 41.00 & 41.00 & 35.00 & 47.00 & 45.00 & 41.50 & 40.50 & 35.50 \\
llama-4-scout\_3-flash & 53.00 & 54.00 & 55.50 & 48.00 & 62.00 & 56.50 & 54.50 & 53.50 & 52.50 \\
ministral-14b\_2.5-flash & 40.50 & 39.00 & 43.50 & 36.50 & 42.50 & 45.50 & 42.50 & 39.00 & 37.00 \\
ministral-14b\_3-flash & 54.50 & 56.00 & 54.50 & 47.00 & 61.00 & 57.00 & 57.50 & 53.00 & 51.00 \\
mistral-small\_2.5-flash & 38.50 & 40.50 & 41.00 & 34.00 & 40.50 & 42.00 & 44.00 & 37.50 & 37.50 \\
mistral-small\_3-flash & 51.00 & 54.50 & 47.00 & 42.50 & 61.00 & 56.50 & 57.50 & 46.00 & 50.00 \\
    \hline
mean & 49.30 & 49.00 & 47.60 & 42.50 & 53.00 & 49.95 & 50.60 & 46.45 & 45.90 \\
    \hline
std & 7.98 & 8.58 & 6.99 & 7.26 & 10.23 & 6.58 & 8.63 & 8.17 & 8.64 \\
  \hline
  \end{tabular}
  \end{adjustbox}

\end{table}

\begin{table}[H]
  \caption{Strong-agent influence (\%) on SuperGPQA in the Weak-Weak-Strong order, reported by model combination.}
  \label{tab:supergpqa_influence_forward}
  \centering
  \begin{adjustbox}{max width=\textwidth}
  \begin{tabular}{lccccccccc}
    & & \multicolumn{4}{c}{Agreeableness} & \multicolumn{4}{c}{Extraversion} \\
    \cline{3-6} \cline{7-10}
    \textbf{Model combination} & \textbf{\textit{No-personality}} & \textbf{\textit{ah\_ah\_n}} & \textbf{\textit{al\_al\_n}} & \textbf{\textit{n\_n\_ah}} & \textbf{\textit{n\_n\_al}} & \textbf{\textit{eh\_eh\_n}} & \textbf{\textit{el\_el\_n}} & \textbf{\textit{n\_n\_eh}} & \textbf{\textit{n\_n\_el}} \\
    \hline
2.5-flash-lite\_2.5-flash & 21.95 & 23.40 & 27.55 & 20.93 & 29.63 & 26.67 & 24.74 & 20.45 & 22.47 \\
2.5-flash-lite\_3-flash & 48.48 & 53.15 & 50.00 & 36.79 & 55.34 & 47.17 & 54.13 & 39.13 & 39.81 \\
4o-mini\_2.5-flash & 50.50 & 50.51 & 31.43 & 29.09 & 57.84 & 54.00 & 52.00 & 39.42 & 39.64 \\
4o-mini\_3-flash & 69.60 & 72.58 & 54.03 & 53.28 & 75.83 & 72.80 & 73.39 & 67.44 & 71.90 \\
llama-4-scout\_2.5-flash & 40.59 & 39.60 & 35.05 & 17.76 & 54.64 & 41.18 & 43.62 & 26.55 & 26.55 \\
llama-4-scout\_3-flash & 49.56 & 62.28 & 55.26 & 43.24 & 69.09 & 57.26 & 57.14 & 56.41 & 53.85 \\
ministral-14b\_2.5-flash & 30.34 & 34.26 & 26.60 & 18.56 & 42.86 & 40.00 & 40.78 & 31.87 & 23.26 \\
ministral-14b\_3-flash & 52.50 & 60.66 & 59.65 & 38.52 & 67.59 & 60.00 & 60.17 & 50.00 & 49.12 \\
mistral-small\_2.5-flash & 33.33 & 38.64 & 33.71 & 25.27 & 48.28 & 47.96 & 46.91 & 33.72 & 25.81 \\
mistral-small\_3-flash & 55.45 & 54.29 & 44.44 & 39.45 & 67.24 & 62.39 & 63.21 & 52.17 & 47.71 \\
    \hline
mean & 45.23 & 48.94 & 41.77 & 32.29 & 56.83 & 50.94 & 51.61 & 41.72 & 40.01 \\
    \hline
std & 13.88 & 14.85 & 12.37 & 11.83 & 13.95 & 13.20 & 13.52 & 14.57 & 16.06 \\
  \hline
  \end{tabular}
  \end{adjustbox}

\end{table}

\begin{table}[H]
  \caption{Final debate accuracy (\%) on AQuA-RAT in the Weak-Weak-Strong order, reported by model combination.}
  \label{tab:aqua_rat_performance_forward}
  \centering
  \begin{adjustbox}{max width=\textwidth}
  \begin{tabular}{lccccccccc}
    & & \multicolumn{4}{c}{Agreeableness} & \multicolumn{4}{c}{Extraversion} \\
    \cline{3-6} \cline{7-10}
    \textbf{Model combination} & \textbf{\textit{No-personality}} & \textbf{\textit{ah\_ah\_n}} & \textbf{\textit{al\_al\_n}} & \textbf{\textit{n\_n\_ah}} & \textbf{\textit{n\_n\_al}} & \textbf{\textit{eh\_eh\_n}} & \textbf{\textit{el\_el\_n}} & \textbf{\textit{n\_n\_eh}} & \textbf{\textit{n\_n\_el}} \\
    \hline
2.5-flash-lite\_2.5-flash & 85.00 & 86.00 & 87.50 & 85.00 & 87.00 & 83.50 & 82.00 & 86.00 & 85.50 \\
2.5-flash-lite\_3-flash & 87.00 & 88.00 & 85.00 & 87.50 & 87.00 & 85.00 & 85.00 & 87.00 & 86.50 \\
4o-mini\_2.5-flash & 82.50 & 81.00 & 74.50 & 77.00 & 79.00 & 82.00 & 83.00 & 81.50 & 80.50 \\
4o-mini\_3-flash & 86.50 & 85.50 & 74.50 & 84.00 & 87.00 & 87.50 & 85.00 & 80.50 & 85.00 \\
llama-4-scout\_2.5-flash & 78.50 & 79.50 & 78.50 & 78.50 & 80.00 & 83.00 & 83.00 & 83.00 & 79.50 \\
llama-4-scout\_3-flash & 82.00 & 82.00 & 85.50 & 81.00 & 86.50 & 84.50 & 82.00 & 83.50 & 83.50 \\
ministral-14b\_2.5-flash & 84.50 & 79.50 & 81.00 & 78.50 & 79.00 & 80.50 & 82.50 & 80.50 & 79.00 \\
ministral-14b\_3-flash & 83.50 & 84.50 & 83.50 & 80.50 & 85.00 & 83.50 & 85.50 & 83.00 & 83.00 \\
mistral-small\_2.5-flash & 85.00 & 81.50 & 82.00 & 80.00 & 81.50 & 82.50 & 84.50 & 79.00 & 81.50 \\
mistral-small\_3-flash & 85.00 & 85.50 & 83.00 & 82.00 & 83.00 & 83.00 & 87.00 & 82.50 & 83.00 \\
    \hline
mean & 83.95 & 83.30 & 81.50 & 81.40 & 83.50 & 83.50 & 83.95 & 82.65 & 82.70 \\
    \hline
std & 2.48 & 2.97 & 4.45 & 3.27 & 3.42 & 1.89 & 1.69 & 2.47 & 2.55 \\
  \hline
  \end{tabular}
  \end{adjustbox}

\end{table}

\begin{table}[H]
  \caption{Strong-agent influence (\%) on AQuA-RAT in the Weak-Weak-Strong order, reported by model combination.}
  \label{tab:aqua_rat_influence_forward}
  \centering
  \begin{adjustbox}{max width=\textwidth}
  \begin{tabular}{lccccccccc}
    & & \multicolumn{4}{c}{Agreeableness} & \multicolumn{4}{c}{Extraversion} \\
    \cline{3-6} \cline{7-10}
    \textbf{Model combination} & \textbf{\textit{No-personality}} & \textbf{\textit{ah\_ah\_n}} & \textbf{\textit{al\_al\_n}} & \textbf{\textit{n\_n\_ah}} & \textbf{\textit{n\_n\_al}} & \textbf{\textit{eh\_eh\_n}} & \textbf{\textit{el\_el\_n}} & \textbf{\textit{n\_n\_eh}} & \textbf{\textit{n\_n\_el}} \\
    \hline
2.5-flash-lite\_2.5-flash & 63.01 & 72.73 & 69.33 & 60.81 & 71.43 & 71.79 & 65.85 & 66.67 & 64.29 \\
2.5-flash-lite\_3-flash & 85.88 & 92.47 & 84.71 & 84.34 & 84.88 & 85.71 & 84.62 & 84.15 & 84.71 \\
4o-mini\_2.5-flash & 75.53 & 74.47 & 62.86 & 66.67 & 77.08 & 77.66 & 75.00 & 70.71 & 77.91 \\
4o-mini\_3-flash & 87.83 & 88.39 & 68.91 & 84.35 & 91.15 & 90.52 & 88.70 & 85.32 & 85.34 \\
llama-4-scout\_2.5-flash & 61.18 & 70.65 & 69.89 & 59.77 & 66.30 & 70.45 & 71.76 & 63.16 & 66.67 \\
llama-4-scout\_3-flash & 80.00 & 89.90 & 86.79 & 79.35 & 83.67 & 85.29 & 83.00 & 79.38 & 76.00 \\
ministral-14b\_2.5-flash & 71.05 & 69.62 & 73.49 & 60.92 & 67.95 & 79.49 & 74.36 & 62.34 & 60.76 \\
ministral-14b\_3-flash & 75.51 & 81.00 & 84.62 & 75.28 & 87.10 & 85.44 & 83.02 & 77.66 & 82.35 \\
mistral-small\_2.5-flash & 61.40 & 67.92 & 59.62 & 54.72 & 58.82 & 56.86 & 60.42 & 50.00 & 55.77 \\
mistral-small\_3-flash & 71.83 & 81.58 & 71.43 & 79.37 & 84.38 & 82.09 & 81.08 & 80.65 & 74.63 \\
    \hline
mean & 73.32 & 78.87 & 73.16 & 70.56 & 77.28 & 78.53 & 76.78 & 72.00 & 72.84 \\
    \hline
std & 9.58 & 9.06 & 9.35 & 11.19 & 10.65 & 9.91 & 8.95 & 11.41 & 10.41 \\
  \hline
  \end{tabular}
  \end{adjustbox}

\end{table}

\end{document}